\documentclass[twoside,11pt]{article}

\usepackage{jmlr2e}

\usepackage{amsmath}
\usepackage{amsfonts}
\usepackage{xr}
\usepackage{accents}
\usepackage{enumitem}

\newcommand{\half}{\frac{1}{2}}
\newcommand{\RR}{\mathbb{R}}

\usepackage{caption} 

\firstpageno{1}

\begin{document}

\title{Studying Generalization Through Data Averaging}

\author{\name Carlos A. Gomez-Uribe \email cgomezuribe@apple.com 
 \\
    \addr Apple}

\editor{TBD}

\maketitle

\begin{abstract}
The generalization of machine learning models has a complex dependence on the data, model and learning algorithm. We study train and test performance, as well as the generalization gap given by the mean of their difference over different data set samples to understand their ``typical" behavior. We derive an expression for the gap as a function of the covariance between the model parameter distribution and the train loss, and another expression for the average test performance, showing test generalization only depends on data-averaged parameter distribution and the data-averaged loss. We show that for a large class of model parameter distributions a modified generalization gap is always non-negative. By specializing further to parameter distributions produced by stochastic gradient descent (SGD), along with a few approximations and modeling considerations, we are able to predict some aspects about how the generalization gap and model train and test performance vary as a function of SGD noise. We evaluate  these predictions empirically on the Cifar10 classification task based on a ResNet architecture.
\end{abstract}


\section{Introduction}

Statistical mechanics studies systems that despite their complexity and richness of possible behaviors, can give rise to simple, interesting and useful relationships under appropriate conditions that are captured by relatively simple equations, e.g., the Arrhenius equation that describes the dependence of chemical reaction rates on temperature and other characteristics. There, the back-and-forth between theory and experiments has often resulted in the deepest understanding. Can we achieve a similar state of affairs in machine learning theory that yields simple, interesting and useful predictions that match experiments about model performance despite the complex ways in which data, model and algorithms interact to determine generalization in today's applications? There is some empirical evidence that simple patterns can indeed emerge in today's applications, e.g., \cite{hestness2017deep, bansal2022data} show that test performance can be well described by a power law function of the training set size. We motivate this paper with other simple empirical patterns from a single modern application described shortly that warrant a theoretical explanation. The main goal of the theory we develop here is understanding as much as possible about these experimental behaviors.

\begin{figure}[htbp]
\centering
\textbf{SGD with Momentum and Cosine Learning Schedule} \par 
\includegraphics[width=0.925\columnwidth]{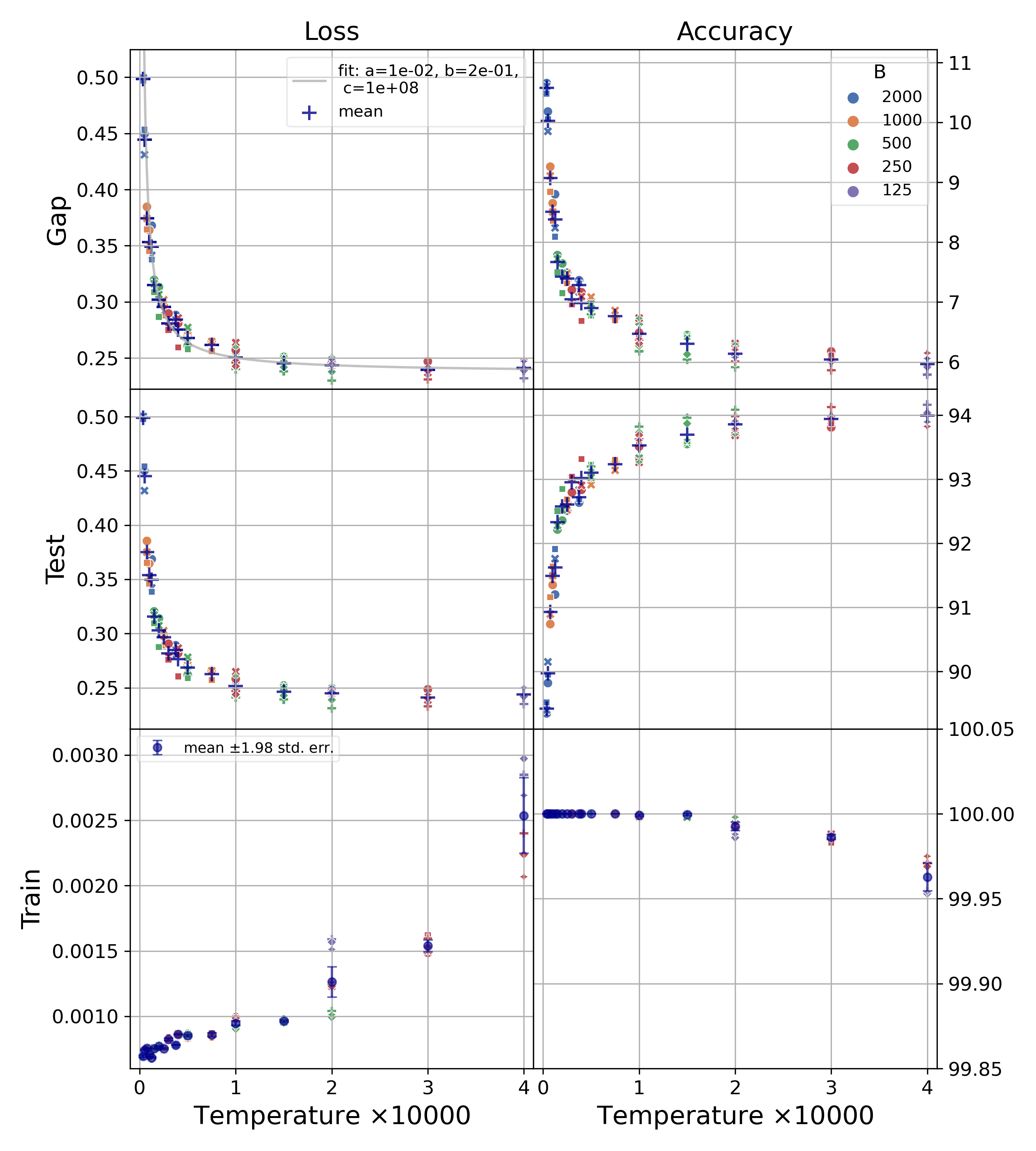}
\caption{ResNet with 2,796,202 parameters trained on Cifar 10 data via SGD with momentum of 0.9 and cosine learning schedule. The x-axis is the SGD temperature, defined as the ratio of the initial learning rate and the batch size. Each data point is a different run started by randomly partitioning the 60k examples in the data set into a training set of 50k examples and a test set with the 10k remaining ones. Different marker types correspond to different random seeds, and there are at least three for each value of $T$ in the plot. The left columns shows the gap, test and train loss. The right column shows the accuracy gap, and the test and train accuracies. The gray line shows the fit of the gap to Eq. \ref{eq:GapVsT} after re-scaling $T$ by multiplying it by $1e4$. So in terms of the raw $T$ the parameter values for $a$ and $c$ need to be divided by $1e4.$}
\label{fig:SGDMomentumCosineK32DataSplitting}
\end{figure}

\begin{figure}[htbp]
\centering
\textbf{SGD with Momentum and Cosine Learning Schedule} \par 
\includegraphics[width=0.925\columnwidth]{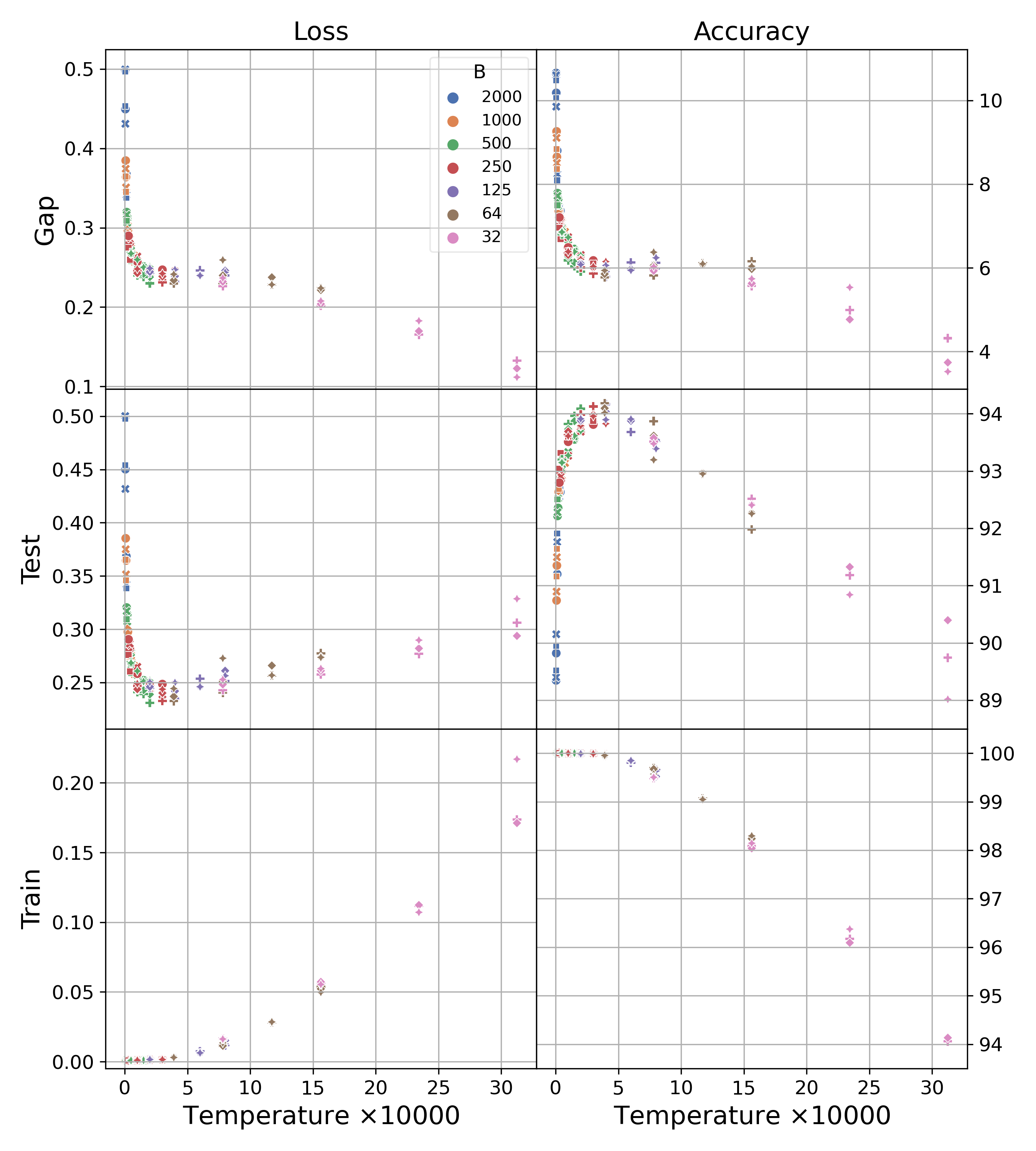}
\caption{Same setup as Fig. \ref{fig:SGDMomentumCosineK32DataSplitting} but for a broader range of temperatures. Different marker types correspond to different random seeds.}
\label{fig:SGDMomentumCosineK32DataSplittingAllTemps}
\end{figure}

Figure \ref{fig:SGDMomentumCosineK32DataSplitting} shows the generalization gap, the gap in accuracy, and the training set and testset losses and accuracies of the running application example in this work: classifying the images of the Cifar10 data set into their 10 categories by training a ResNet model with a variant of the SGD algorithm until convergence. The loss is the typical cross-entropy loss, and the accuracy is the percent of correctly classified examples in the data set (choosing the category for each image that the model finds to be the most likely). The x-axis is the ratio of the learning rate and SGD mini-batch size, which we call the SGD temperature and denote it by $T.$ Each datapoint corresponds to a different experiment run, and different runs differ in some combination of the learning rate, mini-batch size, and random seed used. The random seed also determines the training set and test set through random sampling described next. Cifar10 comes already split into 50,000 training images and 10,000 test images. For each random seed we merge the original training and test sets into a single data set with 60,000 images, reshuffle the images, and re-divide the data into a 50,000 training set and the remaining 10,000 test images. For every value of $T$ we have multiple datapoints, so we also show the mean in dark blue. Note how well the different datapoints seem to align along a single function of $T$ for all metrics. E.g., the average generalization gap which we denote by $\overline{\langle \Delta \rangle}$ and appears in dark blue on the top left plot is well described by the simple function 
\begin{equation}
    \overline{\langle \Delta \rangle} = \biggr(\frac{a}{T}+b \biggr)e^{-\frac{T}{c}} \approx \frac{a}{T}+b, \label{eq:GapVsT}
\end{equation}
where $a,$ $b$ and $c$ are positive constants fit to the data, and where the last approximate equality holds when $c \gg T$ and seems valid in Fig. \ref{fig:SGDMomentumCosineK32DataSplitting} (see gray line and legend). For the range of temperatures shown, although the train loss increases with temperature, it is 2 orders of magnitude smaller than the test loss, so test performance here essentially equals the gap and improves with higher temperatures, but only up to a point. Figure \ref{fig:SGDMomentumCosineK32DataSplittingAllTemps} shows analogous experiments but for a broader range of temperatures, showing that further temperature increases now worsen the test loss and test accuracy. So there is an optimal positive SGD temperature that maximizes test performance. The gap at higher temperatures has a second decrease that is not captured by Eq. \ref{eq:GapVsT}. Similarly, all or nearly all training examples are perfectly classified for the lower temperatures in Fig. \ref{fig:SGDMomentumCosineK32DataSplitting}, so the training accuracy is essentially 100\% in all these experiments, and starts degrading only at the higher temperatures of Fig. \ref{fig:SGDMomentumCosineK32DataSplittingAllTemps}. So train performance only worsens with increased temperature. The approximate theory we develop in this paper aims to explain these experimental results, and make useful practical predictions: why do results line up so cleanly as a function of temperature despite using different batch sizes, learning rates, randomly sampled data sets, and random seeds? Is the gap always positive, and how does this follow mathematically? Can we derive the relations that the gap, the train loss, and test performance follow as a function of temperature? Why does train performance worsen with increased temperature, while test performance first improves and then degrades as temperature increases? Can we predict how these curves would change if we used more or less data, increase the number of model parameters, change the SGD learning rate or batch size?  Although the theory we outline ultimately fails to answer most of these questions fully, it does provide partial answers and offers suggestions for further work, so we hope others can improve upon it until such questions are more clearly answered. Section \ref{sec:mainResults} introduces our setup and notation, and states our main results which include both exact ones that apply to a broad class of learning algorithms, as well as approximate ones, most of them specific to SGD.

\subsection{Contributions}
Our contributions include the following. We derive a new and exact expression for the average gap as an integral over the covariance between the train loss and the distribution of model parameters that arises as data sets are sampled, and obtain a new upper bound for it that depends on the variance across data sets of the parameter distributions and of the loss.  We similarly derive a new expression for the average test performance as a function of an appropriately data-averaged parameter distribution and the data-averaged loss function. We then generalize an earlier result in \cite{seung1992statistical} to a broader class of parameter distributions that shows the positivity of a modified generalization gap. Next, we specialize to situations where SGD is the learning algorithm, and through a series of approximations starting with the steady-state solution of a diffusion process approximation of SGD, we obtain three main predictions: train performance worsens with increased SGD noise, test performance is optimized at an intermediate noise level, and the generalization gap is a simple function of SGD noise specified in Eq. \ref{eq:GapVsT}. Section \ref{sec:ExperimentalResults} evaluates these predictions empirically, describes our experiments in more detail, and provides additional experimental results obtained by using other variants of SGD, as well as a different way to sample data sets that is closer to the theory we develop but leads to experiment results with higher variance. \cite{hardt2016train} and related work also study the generalization gap averaged over data sets and model parameters for SGD through the concept of algorithm stability, obtaining average gap upper bounds that grow linearly with training time; these become vacuous at steady state, unlike ours. Their gap upper bounds also decrease inversely with training set size, and do not apply close to the interpolation regime for large models. Our SGD results do not have this limitation, but they have different ones: only applying to the SGD steady state, and being approximate in nature. \cite{seung1992statistical} studies the generalization gap and related quantities averaging over data sets and model parameters too, but for an algorithm different from SGD (similar to full gradient descent plus constant isotropic noise), and with a focus on understanding how training set size impacts learning. When we make our results algorithm specific, we do so for SGD at steady state, with the goal of understanding how the strength of SGD noise impacts generalization. We connect our work more broadly to previous literature in Section \ref{sec:Discussion}, where we also suggest future work, and conclude with a series of Appendices with additional details. A practical implication of our results is that when learning a model via SGD in order to optimize test performance, a  broad exploration of temperatures that avoids very small temperatures can make a significant difference.

\section{Main Results}
\label{sec:mainResults}
\subsection{Preliminaries}
Let $\mathcal{D}=\{x_\ell, x_e\}$ denote the data, where the subscripts $\ell$ and $e$ correspond to the training (or $\ell$earning) and the test (or $e$valuation) data sets. We assume $x_\ell$ consists of $n_\ell$ i.i.d. samples $x_i^{(\ell)}$ from some distribution $r_\ell(x)$, and similarly for the test set, sampled from distribution $r_e(x)$ that may or may not be the same as $r_\ell(x).$ The distribution of the training set and test set are then, respectively, $p(x_\ell) = \prod_{i=1}^{n_\ell} r_\ell( x_i^{(\ell)})$ and $p(x_e) = \prod_{i=1}^{n_e} r_e( x_i^{(e)})$ . We assume that $r_\ell(x)$ and $r_e(x)$ are known, so that the training and test set are independent. This independence between training and test sets will be central to our results. We denote  the distribution that generates data set samples (of the fixed sizes $n_\ell$ and $n_e$) by $p(\mathcal{D})=p(x_\ell)p(x_e)$. Assume we learn the parameters $\theta \in \RR^p$ of a model by processing $x_\ell$ to produce a distribution $\rho(\theta| x_\ell)$ of model parameters through a stochastic algorithm such as SGD. We consider two sources of randomness: the sampling of $\mathcal{D}$, and the sampling of model parameters from $\rho(\theta|x_\ell)$. We assume we are given a loss function $l(\theta, x) \geq 0$ that is typically the negative log likelihood $-\log p(x|\theta)$ and is used to evaluate model performance and to train the model. We call
\begin{align*}
    U_\ell(\theta, x_\ell) =  \frac{1}{n_\ell} \sum_{i=1}^{n_\ell} l(\theta, x_i^{(\ell)}), \quad &
    U_e(\theta, x_e) =  \frac{1}{n_e} \sum_{i=1}^{n_e} l(\theta, x_i^{(e)})
\end{align*}
the train set and test set losses, and the population averages
$   \overline U_\ell =  E_{x \sim r_\ell}[U_\ell(\theta, x)],$ and $\overline U_e =  E_{x \sim r_e}[U_e(\theta, x)]$
the train and test losses (i.e., no ``set"). We want to understand the train set and test set losses averaged over training outcomes and over data set realizations.  Next, we introduce a compact notation for the main kinds of averages and related quantities we encounter.

Let $f(\theta, \mathcal{D})$ and $h(\theta, \mathcal{D})$ be functions of two random variables $\theta$ and $\mathcal{D}$ with joint distribution $p(\theta, \mathcal{D}) = p(\mathcal{D})\rho(\theta|x_\ell).$ The following notation for key averages is inspired by statistical mechanics and is used throughout:
\begin{align*}
    \langle f \rangle  \triangleq  \int f(\theta, \mathcal{D})\rho(\theta|x_\ell)d\theta = E_{\theta|\mathcal{D}}[f], \quad & 
    \overline f \triangleq  \int f(\theta, \mathcal{D}) p(\mathcal{D})d\mathcal{D} = E_{\mathcal{D}}[f],  \\
    \overline{\langle f \rangle} \triangleq  E_{\theta, \mathcal{D}}[f] = E_\mathcal{D} \big[ E_{\theta| \mathcal{D}}[f]\big], \quad &  \langle \langle  f \rangle \rangle \triangleq  \int \overline \rho(\theta) f(\theta, \mathcal{D}) d\theta, \text{ where}\\ 
     \overline \rho (\theta) \triangleq  \int p(\mathcal{D})\rho(\theta|x_\ell)d\mathcal{D}, \quad & \text{Cov}_\mathcal{D} \big(f, h \big) \triangleq \overline{fh} - \overline{f} \ \overline{h}.
\end{align*}
Note that $\langle f \rangle$ and $\langle \langle  f \rangle \rangle$ are both functions of $\mathcal{D};$ $\overline f$, $\overline h,$ and $\overline{ fh}$ are functions of $\theta$, just like the covariance between $f$ and $h,$ and $\overline{\langle f \rangle}$ is a constant. Sometimes we may explicitly write the arguments of these kinds of functions for clarity, e.g., $\overline f  (\theta)$.  The notation above is important to make what follows clear and avoid a more convoluted alternative.  E.g., note that in general $ \langle \langle  \overline f \rangle \rangle \neq \overline{\langle f \rangle}$ since the latter is the total average of $f$ while the former first averages $\rho$ and $f$  separately over the data, and only then averages over $\theta.$ Sometimes we need the covariance of two quantities across realizations of random variables different from $\mathcal{D}.$ In those cases we indicate in the subscript of the covariance the random variable used to compute the expectations, e.g., $\text{Cov}_{x_\ell}\big(f, h\big)$ is the mean under $p(x_\ell)$ of $f(\theta, \mathcal{D})h(\theta, \mathcal{D})$ minus the product of the corresponding means under $p(x_\ell).$
\subsection{General Results}
In this section we cover results that are not specific to SGD, and we make a distinction between exact and approximate results.
\subsubsection{Exact Results}
First we state a preliminary result:
\begin{align}
   \overline{\langle f \rangle} = & \int p(\mathcal{D}) \rho(\theta|x_\ell) f(\theta, \mathcal{D}) d\mathcal{D}d\theta = \int \overline{\rho(\theta|x_\ell) f(\theta, \mathcal{D})} d\theta \nonumber \\
   = & \int \text{Cov}_\mathcal{D}\big(\rho, f \big) d\theta + \int \overline \rho (\theta) \overline f (\theta) d\theta = \int \text{Cov}_\mathcal{D}\big(\rho, f \big) d\theta + \langle \langle \overline f \rangle \rangle. \label{eq:totalAvgF}
\end{align}
This decomposes the total expectation of any function $f(\theta, \mathcal{D})$ into an integral over the covariance of $\rho$ and $f$ under $p(\mathcal{D})$, and the expectation $\langle \langle \overline f \rangle \rangle$ of the data-averaged $f$ under the data-averaged $\rho.$ In our setting where the train and test sets are independent, any function of the test set but not of the train set, e.g., the test loss $U_e(\theta, x_e),$ has zero covariance with $\rho$ so the integral in the last equality above vanishes and $\overline{\langle f \rangle} = \langle \langle \overline{f} \rangle \rangle.$ With our new notation, we restate our goal as understanding the behavior of $\overline{\langle U_e \rangle}$ and $\overline{\langle U_\ell \rangle}$. We also let $\Delta(\theta, \mathcal{D})=U_e - U_\ell$ be the gap, and we also want to understand its total average $\overline{\langle \Delta \rangle}$. Applying Eq. \ref{eq:totalAvgF} to the train set and test set losses, and to the gap we find that

\begin{align}
    \overline{\langle U_\ell \rangle} = & \int \text{Cov}_\mathcal{D}\big(\rho, U_\ell \big) d\theta + \langle \langle \overline U_\ell \rangle \rangle, \label{eq:train_loss_bar_brak} \\
    \overline{\langle U_e \rangle} = & \int \text{Cov}_\mathcal{D}\big(\rho, U_e \big) d\theta +\langle \langle \overline U_e \rangle \rangle  =\langle \langle \overline U_e \rangle \rangle , \label{eq:test_loss_bar_brak}\\
    \overline{\langle \Delta \rangle} = & \int \text{Cov}_\mathcal{D}\big(\rho, U_e - U_\ell \big) d\theta + \langle \langle \overline \Delta \rangle \rangle  \nonumber \\
    = & -  \int \text{Cov}_\mathcal{D}\big(\rho,  U_\ell \big) d\theta +  \langle \langle \overline \Delta \rangle \rangle, \label{eq:gap_bar_brak}
\end{align}
where we used the fact that $\text{Cov}_\mathcal{D}\big(\rho, U_e \big)=0$ for all $\theta,$ since $\rho$ depends only on $x_\ell,$ and $U_e$ only on $x_e,$ and $x_\ell$ and $x_e$ are independent by assumption. The previous three equations are our first set of results. They can be re-arranged for the typical interpretation of test performance being equal to empirical risk minimization plus generalization gap:
\begin{align}
    \overline{\langle U_e \rangle} = \overline{\langle U_\ell \rangle} + \biggr(  \langle \langle \overline \Delta \rangle \rangle -  \int \text{Cov}_\mathcal{D}\big(\rho,  U_\ell \big) d\theta \biggr). \nonumber
\end{align}
This equation suggests the common approach of breaking up the task of achieving good test performance into minimization of the training loss (or epmirical risk), and then hoping that the gap (in parenthesis) can be designed to be small, e.g., through an appropriate choice of model architecture or learning recipe. We hypothesize it may also be useful to directly consider the test performance and how to optimize it, and Eq. \ref{eq:test_loss_bar_brak} tells us we then need to understand  $\overline{ \rho }(\theta)$ and $\overline{ U_e }(\theta).$ However, understanding the gap may turn out to be simpler, and in the so-called interpolation regime, commonly seen in practice today where the train loss is very close to zero, the test loss is essentially equal to the gap, so understanding the gap becomes equivalent to understanding test performance.

When there is no distribution shift and $r_\ell(x)=r_e(x),$ our results simplify since $\langle \langle \overline U_e \rangle \rangle = \langle \langle \overline U_\ell \rangle \rangle = \langle \langle \overline U \rangle \rangle,$ so that $\langle \langle \overline \Delta \rangle \rangle = 0$, and in particular the gap is simply the integral of the data set-sampling-induced covariance between the parameter distribution and the train loss:
\begin{align}
    \overline{\langle \Delta \rangle} =  -  \int \text{Cov}_{x_\ell}\big(\rho,  U_\ell \big) d\theta. \label{eq:gap_bar_brak2}
\end{align}
The gap is then a measure of the covariance between the parameter distribution and the train loss as you sample different training sets, a seemingly different mathematical object than the difference between test and train performance that opens up new possibilities for its analysis. In the rest of the paper, we assume no distribution shift to minimize scope. We study Eq. \ref{eq:gap_bar_brak2} further through special cases and approximations later, but can already obtain an upper bound for it. Recall that for any two random variables, the magnitude of their covariance is upper bounded by the product of their standard deviations. In addition, since $\text{Var}_{x_\ell}\big( U_\ell \big) = \frac{1}{n_\ell} \sigma_l^2(\theta)$ where $\sigma_l^2(\theta)$ is the variance over $r_\ell(x)$ of the loss $\ell(\theta, x),$ we find that
\begin{align}
   \big| \overline{\langle \Delta \rangle} \big| \leq  \frac{1}{\sqrt{n_\ell}} \int \sigma_\rho(\theta)  \sigma_l(\theta) d\theta. \label{eq:gap_upperbnd}
\end{align}
Here $\sigma_\rho(\theta)$ is the standard deviation of $\rho(\theta|x_\ell)$ under training set sampling. So, e.g., algorithms that produce parameter distributions that vary little across data set sampling in parameter regions where the loss function varies highly across data set samples, or models with losses that vary little across data set samples for parameter values where the likelihood of parameter varies highly across data set samples, should yield small gaps. We expect $\sigma^2_\rho(\theta)$ to have some dependence on $n_\ell,$ so the bound in Eq. \ref{eq:gap_upperbnd} is only partially explicit as a function of $n_\ell.$ Depending on the training method, $\sigma_\rho(\theta)$ can bring up another $1/\sqrt{n_\ell}$ factor, e.g., when $\rho(\theta|x_\ell) \propto e^{-\frac{U_\ell(\theta, x_\ell)}{T}},$ as considered in \cite{seung1992statistical} and \cite{gelfand1991recursive}. However in other situations, such as when SGD is used for training and as $n_\ell$ approaches the interpolation regime, $\sigma_\rho(\theta)$ might actually grow with $n_\ell$; this could be studied in future work. The results presented also motivate seeking a better understanding of $\text{Cov}_{x_\ell}\big(\rho,  U_\ell \big),$  
$\sigma^2_\rho(\theta),$ and $\sigma^2_\ell(\theta).$ Some aspects of the latter are straightforward to explore experimentally and pursued later.

A deceivingly simple question is whether the average gap is positive. We struggle to prove it exactly, but provide related results, the first of which needs an additional assumption: the parameter distribution follows the Boltzmann-like form:
\begin{equation}
    \rho(\theta|x_\ell) = \frac{1}{Z} e^{-g(\theta, x_\ell)}, \text{ where } Z = \int e^{-g(\theta, x_\ell)} d\theta \label{eq:Boltzmann}
\end{equation}
is called the partition function, assumed finite, and where $g(\theta, x_\ell)$ is called the effective potential. We interpret the above distribution as a soft minimization of $g(\theta, x_\ell)$ over $\theta,$ since regions where the effective potential is large have little probability density, and vice-versa. We will see later that SGD can produce a distribution of this form for a specific effective potential that is related to, but different from, the train loss. \cite{seung1992statistical}'s study of related questions also assumes a specific instance of the above, with the effective potential being proportional to the train loss. For any distribution of the form in Eq. \ref{eq:Boltzmann} there is a positive effective potential gap, i.e.,
\begin{equation}
    \overline{\langle g \rangle} \leq \langle \langle \overline{g} \rangle \rangle. \label{eq:effPotGap}
\end{equation}
The quantity on the left of Eq. \ref{eq:effPotGap} is the (total average of the) training effective potential, and that on the right is the (total average of the) test effective potential. When the effective potential is sufficiently similar to the model loss function, Eq. \ref{eq:effPotGap} directly implies a non-negative loss gap.  To prove Eq. \ref{eq:effPotGap} we adapt the argument in Appendix A of \cite{seung1992statistical} to our more general setup with an arbitrary effective potential. Consider a parameter distribution $q(\theta)$ that interpolates between $\rho(\theta|x_\ell)$ and some mean-field data-independent distribution as follows:
\begin{equation*}
    q(\theta) = \frac{1}{Z_q}e^{-c \delta g(\theta, x_\ell) - \overline{g}(\theta)},
\end{equation*}
where $c \geq 0$  is a scalar, $Z_q$ is the partition function, and where we define the residual potential $\delta g(\theta, x_\ell) = g(\theta, x_\ell) - \overline{g}(\theta)$. So $\overline{\delta g}=0$ for all $\theta$ by design. Setting $c=1$ makes $q(\theta) = \rho(\theta|x_\ell),$ while setting $c=0$ makes $q(\theta)$ have a data-independent potential $\overline{g}(\theta).$ Direct calculations then yield
    $\partial_{c} \log Z_q = - \langle \delta g \rangle_q,$ and
    $\partial_{c}^2 \log Z_q =  \text{Var}_q \big(\delta g \big) = - \partial_{c}\langle \delta g \rangle_q,$
where the angle-brackets above are averages relative to $q(\theta).$ The last equation means that $ \partial_{c}\langle \delta g \rangle_q \leq 0,$ and in particular, that
$\langle  \delta g \rangle_{q, c > 0} \leq \langle \delta g \rangle_{q, c = 0}.$
I.e., the average of the residual potential is largest when $c=0,$ which is when $q(\theta)$ is data independent. Next, we average the previous expression over data set realizations to get $\overline{\langle  \delta g \rangle}_{q, c > 0} \leq \langle \overline{\delta g} \rangle_{q, c = 0} = 0,$ since we are able to move the averaging inside the integral when $c=0$ because $q(\theta)$ is data-independent, and $\overline{\delta g}=0$ by construction. We finally set $c=1$ on the left average to get $\overline{\langle  \delta g \rangle}=\overline{\langle   g \rangle} - \langle \langle \overline{g} \rangle \rangle  \leq  0,$ where we used  $\overline{\langle \overline{g} \rangle} = \langle \langle \overline{g} \rangle \rangle,$ yielding Eq. \ref{eq:effPotGap}.

\subsubsection{Approximate Results}
Now we seek to make some of expressions more intuitive and explicit through approximations that are hard to justify rigorously but produce useful results, e.g., that match experiments. We first focus on the gap in Eq. \ref{eq:gap_bar_brak2} and approximate the covariance between $\rho$ and $U_\ell$ directly via a log-normal approximation combined with some applications of the delta method in statistics. In Appendix \ref{sec:appDeltaMethod} we pursue a different approximation expected to be valid under less general conditions that is based entirely on the delta method that still yields similar results: relating the gap to the covariance between the effective potential and the loss function. Recall that when $x$ is a scalar Gaussian random variable with mean $\mu$ and variance $\sigma^2,$ the function $y=e^x$ has a log-normal distribution with mean $\mu_y=e^{\mu + \frac{1}{2}\sigma^2}$ and variance $\sigma_y^2=(e^{\sigma^2}-1)\mu_y^2$. Ignoring the stochasticity in $Z$ and assuming $g(\theta, x_\ell)$ is approximately normal with mean and variance $\overline g(\theta)$ and $\sigma^2_g(\theta),$ we can approximate $e^{-g(\theta, x_\ell)}$ as log-normal to find that 
\begin{equation}
    \overline{\rho}(\theta) \approx \frac{1}{\overline Z}e^{-\overline g(\theta) +\half \sigma^2_g(\theta)}, \text{ where }  \overline Z = \int e^{-\overline g(\theta) +\half \sigma^2_g(\theta)} d\theta, \text{ and } \sigma^2_\rho (\theta) \approx \overline{\rho}^2 (e^{\sigma^2_g(\theta)}-1).
    \label{eq:barRhoLogNormal}
\end{equation}
We assume $\overline Z$ is finite. So $\overline{\rho}(\theta)$ appears to simultaneously attempt to minimize the mean-field effective potential $\overline{g}$ while maximizing its variance across data sets. Substituting the approximate expression for $\sigma_\rho^2(\theta)$ in Eq. \ref{eq:gap_upperbnd} the upper bound on the gap becomes
\begin{align}
    \overline{\langle \Delta \rangle} \leq \frac{1}{\sqrt{n_\ell}} \langle \langle \sigma_\ell \sqrt{e^{\sigma^2_g}-1}\rangle \rangle \approx \frac{1}{\sqrt{n_\ell}} \times
    \begin{cases} 
    \langle \langle \sigma_\ell e^{\half\sigma^2_g} \rangle \rangle \text{ when $\sigma_g^2(\theta)  \gg 1$,}
    \\
    \langle \langle \sigma_\ell \sigma_g\rangle \rangle \text{ when $\sigma_g^2(\theta)  \ll 1$.}
    \end{cases}
    \label{eq:gap_upperbnd_approx}
\end{align}
So the gap upper bound grows as the effective potential varies more across different training sets, with a specific functional form that depends on the magnitude of $\sigma_g^2(\theta).$  A similar approximation based on the log-normal distribution can be used on the gap directly, rather than on its upper bound\footnote{We are after $\text{Cov}_{x_\ell} (\rho, U_\ell) = \overline{\rho U_\ell} - \overline{\rho} \overline{U_\ell}.$ We write the first term as $\overline{\frac{1}{Z}e^{-g(\theta, x_\ell) + \ln U_\ell (\theta, x_\ell)}}$ and ignore again the stochasticity of $Z,$ while assuming that $ \ln U_\ell (\theta, x_\ell)$ is approximately Gaussian to apply the log-normal approximation and obtain
\begin{equation*}
    \overline{\rho U_\ell} \approx \frac{1}{\overline{Z}}e^{-\overline g + \half \sigma^2_g +  \overline{ \ln U_\ell} + \half \text{Var}(\ln U_\ell) - \text{Cov}(g, \ln U_\ell) }. 
\end{equation*}
We then use the delta method to obtain $\overline{ \ln U_\ell} \approx \ln \overline U_\ell - \half \frac{\sigma_u^2(\theta)}{\overline{U_\ell}^2(\theta)},$ where $\sigma_u^2(\theta)=\text{Var}_{x_\ell}(U_\ell),$ and $\text{Var}_{x_\ell}(\ln U_\ell) \approx \frac{\sigma_u^2(\theta)}{\overline{U_\ell}^2(\theta)}\big(1-\frac{1}{4}\frac{\sigma_u^2(\theta)}{\overline{U_\ell}^2(\theta)} \big)$. The latter is a variance and must be non-negative, which means the delta method approximation requires $\frac{\sigma_u^2(\theta)}{\overline{U_\ell}^2(\theta)} \leq 4.$  The delta method also yields $\text{Cov}_{x_\ell}(g, \ln U_\ell) \approx \frac{\sigma_{g, u}(\theta)}{U_\ell(\theta, \mu)} \approx \frac{\sigma_{g, u}(\theta)}{\overline U_\ell(\theta)} $. So $
    \overline{\rho U_\ell} \approx \overline{\rho}\overline{ U_\ell}e^{ - \frac{\sigma_{g, u}(\theta)}{\overline U_\ell(\theta)} -\frac{1}{8}\frac{\sigma_u^4(\theta)}{\overline{U_\ell}^4(\theta)}},$ and Eq. \ref{eq:covRhoU} follows. }, to find that
\begin{equation}
    \text{Cov}_{x_\ell} (\rho, U_\ell) \approx -\overline \rho(\theta) \overline U_\ell (\theta) \big(1 - e^{ -  \frac{\sigma_{g, u}(\theta)}{\overline U_\ell(\theta)}  -\frac{1}{8}\frac{\sigma_u^4(\theta)}{\overline{U_\ell}^4(\theta)}}\big), \label{eq:covRhoU}
\end{equation}
which implies
\begin{align}
    \overline{\langle \Delta \rangle} \approx & \langle \langle \overline{U} \big(1 - e^{ -  \frac{\sigma_{g, u}(\theta)}{\overline U(\theta)}  -\frac{1}{8}\frac{\sigma_u^4(\theta)}{\overline{U}^4(\theta)}}\big)\rangle \rangle 
    \approx 
    \begin{cases}
    \langle \langle \overline U_\ell \rangle \rangle, \text{ when $  \frac{\sigma_{g, u}(\theta)}{\overline U(\theta)}  +\frac{1}{8}\frac{\sigma_u^4(\theta)}{\overline{U}^4(\theta)}\gg 1$} \\  
    \langle \langle \sigma_{g,u}\rangle \rangle + \frac{1}{8}\big\langle \big\langle\frac{\sigma_u^4}{\overline{U}^3}\big\rangle \big\rangle, \text{ when $  \frac{\sigma_{g, u}(\theta)}{\overline U(\theta)}  +\frac{1}{8}\frac{\sigma_u^4(\theta)}{\overline{U}^4(\theta)}\ll 1$.}
    \end{cases}\label{eq:gapApproxLog}
\end{align}
 This gap approximation is non-negative, e.g., if $\sigma_{g,u}(\theta)$ is non-negative for all $\theta$. When $  \frac{\sigma_{g, u}(\theta)}{\overline U(\theta)}  +\frac{1}{8}\frac{\sigma_u^4(\theta)}{\overline{U}^4(\theta)}\ll 1$ the above is consistent with the delta method approximation in Eq. \ref{eq:gapApproxDelta} and with the approximate upper bound in Eq. \ref{eq:gap_upperbnd_approx} in situations where $\sigma_g^2(\theta) \ll 1$. On the other hand, when $  \frac{\sigma_{g, u}(\theta)}{\overline U(\theta)}  +\frac{1}{8}\frac{\sigma_u^4(\theta)}{\overline{U}^4(\theta)}\gg 1$ our gap approximation becomes approximately equal to the test performance, so the train performance must be close to zero, i.e., we must be in the interpolation regime. In this setting, and if $\sigma^2_g(\theta) \gg 1$ too,  we can combine Eqs. \ref{eq:gap_upperbnd_approx} and \ref{eq:gapApproxLog} to obtain an approximate upper bound for test performance \begin{align}
     \overline{\langle U_\ell \rangle} \leq \frac{1}{\sqrt{n_\ell}} \langle \langle \sigma_\ell e^{\frac{ \sigma^2_g}{2}}\rangle \rangle. \label{eq:test_perf_upper_approx}
 \end{align}
 We refer to Eq. \ref{eq:gapApproxLog} as the log-normal approximation of the gap, and expect it to be more accurate than that obtained via the delta method in Eq. \ref{eq:gapApproxDelta} for all values of $\sigma_{g,u}(\theta),$ since the latter implicitly assumes that both $\sigma_{g,u}(\theta)$ and $\frac{\sigma_u^4(\theta)}{\overline{U}^3(\theta)}$ are small. But these results are still not explicit enough to produce predictions that can be tested experimentally. A main challenge is that we do not generally know in practice what $\rho(\theta|x_\ell)$ or $g(\theta, x_\ell)$ are in closed form. Specializing to SGD as the training algorithm partially alleviates these challenges, and is pursued next.

\subsection{Stochastic Gradient Descent Results}

We consider the stochastic gradient descent learning algorithm with the following dynamics:
\begin{align}
 \theta_{t+1} &= \theta_t  -\frac{\lambda}{B}\sum_{i=1}^B \partial_\theta U(x_i, \theta_t) - \lambda \alpha \theta_t + \sqrt{\lambda T} \beta w_t, \label{eq:SGDMod} 
\end{align}
where $t$ denotes the training time or number of algorithm updates, $\alpha, \beta$ are scalars, $T=\lambda/B$ is called the temperature, and $w_t$ is zero-mean Gaussian noise in $\RR^p$ with covariance equal to the identity that is independent of everything else. The second term is the loss gradient averaged over $B$ training examples chosen uniformly amongst the training set and with replacement. The number of examples in each update $B$ is called the mini-batch size, and $\lambda$ is called the learning rate. Typically $\lambda \ll 1,$ so $T \ll 1$ also. The third term comes from $\ell_2$ regularization of the train loss, and the last term is isotropic noise. We refer to this version of SGD as plain SGD, because it is often some variant of it that is used in practice, including in some of our experiments. When $\lambda_t$ is allowed to depend on time, the resulting algorithm is SGD with a learning schedule. Most learning schedules decrease the learning rate over the course of training, and we define $T$ in these situations based on the initial learning rate. When we experiment with learning schedules in this work we use the so-called cosine learning schedule $\lambda_t = \lambda \cos( \frac{\pi}{2} \frac{t}{t_f}),$ where $t_f$ is the total number of SGD updates executed during training and $T$ remains defined by $\lambda/B$. The other SGD variant we experiment with is SGD with momentum, aimed at accelerating convergence. Momentum adds another set of state variables reminiscent of the physical variables with the same name:
\begin{align}
    v_{t+1}=\mu v_t + \frac{1}{B}\sum_{i}^B \partial_\theta U(x_i, \theta_t) + \alpha \theta_t & \text{, and } \theta_{t+1} = \theta_t - \lambda v_{t+1} . \label{eq:SGDMomentum}
\end{align}
Here $v_t$ are the momentum variables and $\mu$ a scalar parameter called the momentum coefficient, which we set to a typical value of 0.9 when we use this algorithm. We write SGD with momentum in the form above to match the Pytorch implementation, which we use in our experiments. Figure \ref{fig:SGDMomentumCosineK32DataSplitting} used SGD with momentum and a cosine learning schedule. Because our experiments focus on the dependence with respect to $T,$ we keep $\alpha$ constant at $5e^{-4}.$

To further understand the performance of models trained with SGD we want the distribution of model parameters produced in an explicit-enough form. Unfortunately, we only know a way to obtain an approximation of it for plain SGD, without momentum or a learning schedule. So the theory that follows assumes the plain SGD in Eq. \ref{eq:SGDMod}. Pleasantly, some of the resulting predictions seem to hold empirically for other SGD variants as well, with the exception of SGD with momentum without a learning schedule. To make progress we now approximate the discrete-time stochastic process in Eq. \ref{eq:SGDMod} with a continuous-time diffusion process described by the following Langevin, or stochastic differential equation:

\begin{align}
d \theta = -\lambda \big(\partial_\theta  U_\ell(\theta, x_\ell) + \alpha \theta \big)dt + \sqrt{\lambda T} \big(D(\theta, x_\ell )+ \beta^2 I \big)^{\half}dW, \label{eq:langevinSGD}
\end{align}
where $dW$ denotes the standard Wiener process with identity covariance, and where 
\begin{align*}
    \partial_\theta   U_\ell(\theta,  x_\ell) &= \frac{1}{n_\ell}\sum_j \partial_\theta l_j, \text{  and }
 D(\theta,  x_\ell) =  \frac{1}{n_\ell}\sum_j \partial_\theta l_j \partial_\theta l_j' - \partial_\theta   U_\ell \partial_\theta   U_\ell', \quad \text{and } l_j =  l(\theta, x_j^{(\ell)}). 
\end{align*}
So $D$ is the covariance of the gradients in the training set, which is only invertible when there are fewer parameters than training examples. This diffusion approximation to SGD has appeared in some recent works and its derivation is explained in detail elsewhere, e.g., in \cite{bradley2021can} and briefly in Appendix \ref{sec:diffusionApproximations}. The steady-state distribution of this diffusion process has the Boltzmann form in Eq. \ref{eq:Boltzmann} with 
\begin{equation}
    g(\theta, x_\ell)=\frac{2}{T}v(\theta, x_\ell) + a(\theta, x_\ell), \label{eq:SGDeffPot}
\end{equation}
where 
\begin{align}
     v(\theta, x_\ell) =& \int^\theta \big(D(\theta', x_\ell) + \beta^2 I \big)^{-1}\big(\partial_{\theta'} U_\ell(\theta', x_\ell) + \alpha \theta'\big) \cdot d\theta',\text{ and}\label{eq:v} \\
      a(\theta, x_\ell) =&\int^\theta \big(D(\theta', x_\ell) + \beta^2 I \big)^{-1}\big(\partial_{\theta'} \cdot D(\theta', x_\ell) \big) \cdot d\theta'. \label{eq:a}
\end{align}
We added the last two terms to SGD in Eq. \ref{eq:SGDMod} to obtain the Boltzmann-like distribution just specified regardless of the number of model parameters and training examples. Specifically, we need $\alpha > 0$ and $\beta \neq 0$ for this parameter distribution to hold when there are more model parameters than training examples. Otherwise, with more training examples than model parameters, $\alpha$ and $\beta$ can be zero too without affecting the form of the parameter distribution specified. Eq. \ref{eq:v} conveys the intuition that $v$ is related to $U_\ell$, e.g., if $D(\theta)$ is close to the identity, then $v \propto U_\ell$, and in general we hope that $v$ and $U_\ell$ are positively correlated for different training sets. Appendix \ref{sec:SGDSS} shows a way to check that the above indeed is a steady-state solution of the diffusion process in Eq. \ref{eq:langevinSGD}. There we also show that this diffusion process can be expressed in a rescaled time $t'=\lambda t$ that leaves $T$ and $\alpha$ as the only parameters. This is why we expect that experiments using different batch sizes and learning rates at steady state to only depend on $B$ and $\lambda$ through $T,$ and the reason why we plot our experiments as a function of $T$. We expect the diffusion approximation of SGD to hold for small enough learning rates only, so for small enough temperatures only; see \cite{bradley2021can} for more details. This means that for large enough temperatures we may see that different choices of $B$ and $\lambda$ will not line up well when plotted as a function of $T,$ and we see evidence of this effect in later experiments.

Eq. \ref{eq:SGDeffPot} allows us to specialize some of our results for SGD. First, the positive effective gap in Eq. \ref{eq:effPotGap} now says that
\begin{equation}
    \overline{\langle v \rangle} + \frac{T}{2}\overline{\langle a \rangle} \leq \langle \langle \overline{v} \rangle \rangle + \frac{T}{2} \langle \langle \overline{a} \rangle \rangle. \label{eq:effPotGapSGD}
\end{equation}
For small enough $T$, or when $a$ is small compared to $v,$ the above implies a positive gap for $v(\theta, x),$ and when this function is similar to the loss would imply a positive gap in the loss too. Since $\sigma_{g,u}(\theta) = \frac{2}{T}\sigma_{v,u}(\theta) + \sigma_{a,u}(\theta),$ where the latter functions of $\theta$ are the covariance of $v$ and $a$ with $U_\ell(\theta, x_\ell),$ respectively, Eq. \ref{eq:gapApproxLog}  becomes
\begin{equation}
    \overline{\langle \Delta \rangle} \approx \begin{cases} 
    \langle \langle \overline{ U_\ell }\rangle \rangle \text{, when $  \frac{\sigma_{g, u}(\theta)}{\overline U(\theta)}  +\frac{1}{8}\frac{\sigma_u^4(\theta)}{\overline{U}^4(\theta)}\gg 1$}\\
    \frac{2}{T}\langle \langle \sigma_{v,u}\rangle \rangle + \langle \langle \sigma_{a,u}\rangle \rangle + \frac{1}{8}\big\langle \big\langle\frac{\sigma_u^4}{\overline{U}^3}\big\rangle \big\rangle \text{, when $  \frac{\sigma_{g, u}(\theta)}{\overline U(\theta)}  +\frac{1}{8}\frac{\sigma_u^4(\theta)}{\overline{U}^4(\theta)}\ll 1$.}
    \end{cases} \label{eq:gapApproxLogSGD}
\end{equation}
Note the explicit dependence on $T$ in the latter case: increasing temperature would reduce the gap like $T^{-1}$ ignoring the implicit temperature dependence of the averages. Assuming $\sigma_{v, u}(\theta) > 0,$ we can make  $\sigma_{g, u}(\theta)$ arbitrarily large and positive by making $T$ small enough, leading to being in the interpolation regime. But as $T$ is increased the above does not say whether the model remains in the interpolation regime, just that the gap will follow the approximate form in the lower portion of Eq. \ref{eq:gapApproxLogSGD}. Similarly, since $\sigma^2_g(\theta) = \frac{4}{T^2}\sigma^2_v(\theta)+\sigma^2_a(\theta),$ it can always be made arbitrarily large for small enough $T$ as long as $\sigma^2_v(\theta) > 0,$ so the upper bound in Eq. \ref{eq:gap_upperbnd_approx} becomes
\begin{align}
    \overline{\langle U_\ell \rangle} \approx \overline{\langle \Delta \rangle} \leq  \frac{1}{\sqrt{n_\ell}} \langle \langle \sigma_\ell e^{\frac{2}{T^2}\sigma^2_v + \half \sigma^2_a} \rangle \rangle,
    \label{eq:gap_upperbnd_approx_SGD}
\end{align}
where the left approximate equality assumes $\sigma_{g,u}(\theta)$ is large too. This suggests an exponentially fast deterioration of test performance and the gap as temperature is decreased towards zero. When $\sigma^2_g(\theta) \ll 1,$ which in practice can only happen if  $\sigma^2_v(\theta)$ and $\sigma^2_a(\theta)$ are very small since $T \ll 1,$ we then have the upper bound for the gap
\begin{align}
     \overline{\langle \Delta \rangle} \leq  \frac{2}{ T}\frac{1}{\sqrt{n_\ell}} \big \langle \big \langle \sigma_\ell \sigma_v \sqrt{1 + \frac{T^2}{4}\frac{\sigma^2_a}{\sigma^2_v}} \big \rangle  \big \rangle.
    \label{eq:gap_upperbnd_approx_SGD2}
\end{align}
Eq. \ref{eq:gap_upperbnd_approx_SGD2}  again suggests the average gap upper bound increases as $T$ gets closer to zero. Our experiments in Fig. \ref{fig:SGDMomentumCosineK32DataSplitting} and later figures confirm that both test performance and the gap both deteriorate quickly as $T$ approaches zero from the right. Indeed, the experiments in Figure \ref{fig:SGDMomentumCosineK32DataSplitting}, as well as the form for the gap in Eq. \ref{eq:GapVsT},  were ``inspired" by Eq. \ref{eq:gapApproxLogSGD}. Eq. \ref{eq:GapVsT} follows in three stages: hoping that the gap keeps a similar temperature dependence in the interpolation regime, hoping that the averages in Eq. \ref{eq:gapApproxLogSGD} would have a weak temperature dependence at low temperatures leading to $\frac{a}{T}+b,$ and assuming that the covariances in Eq. $\ref{eq:gapApproxLogSGD}$ go to zero exponentially as the temperature gets very large. The latter is consistent with the fact that no learning happens when the learning rate is large enough, with the train and test loss often going to infinity and producing a gap of zero. This highlights how our analysis falls short; ideally Eq. \ref{eq:GapVsT} would follow mathematically from our treatment rather than just being ``inspired'' by it, and Eq. \ref{eq:gapApproxLogSGD} is as close as our treatment gets to that.

\subsubsection{More on Test and Train Performance}
The log-normal approximation to $\overline{\rho}(\theta)$ in Eq. \ref{eq:barRhoLogNormal} can be studied further to understand the test performance $\langle \langle \overline U \rangle \rangle.$ For SGD that approximation for $\overline{\rho}(\theta)$ becomes
\begin{equation}
    \overline{\rho}(\theta) \approx \frac{1}{\overline Z}e^{-\frac{2}{T}\overline v(\theta) - \overline a(\theta) +\frac{2}{T^2} \sigma^2_v(\theta) + \half \sigma^2_a(\theta) + \sigma_{v,a}(\theta)},  \label{eq:barRhoLogNormalSGD}
\end{equation}
with $\overline Z$ appropiately defined to make $\overline{\rho}(\theta)$ a distribution, and assumed finite.  As $T$ goes to zero the above distribution maximizes $\sigma_v^2(\theta),$ which is not obviously useful in any way. As $T$ is increased $\overline{\rho}(\theta)$ starts to also minimize $\overline{v}(\theta),$ and since we expect this to be similar to  $\overline{U}(\theta),$ that is when SGD appears to target the objective we want. Further increasing $T$ leads to optimizing the non-temperature dependent terms above, which seem unrelated to the test loss objective. So we expect SGD to yield optimal test loss performance at some positive and finite temperature $T.$ This can be argued more precisely as follows. Introducing $\beta=\frac{2}{T}$ for simplicity so that $g(\theta, x_\ell)=\beta v(\theta, x_\ell) + a(\theta, x_\ell)$ direct calculations produce
\begin{align}
    \partial_\beta \ln \overline Z = & \langle \langle -\overline{v} + \beta \sigma^2_v + \sigma_{v, a} \rangle \rangle, \text{ and} \nonumber\\
    \partial_\beta  \overline \rho(\theta) = & \overline \rho(\theta)\big(\langle \langle \overline{v} - \beta \sigma^2_v - \sigma_{v, a} \rangle \rangle - ( \overline{v} - \beta \sigma^2_v - \sigma_{v, a}) \big), \text{ so} \nonumber \\
    \partial_\beta \langle \langle \overline{U} \rangle \rangle =& \int \partial_\beta \overline \rho(\theta) \overline{U}(\theta)d\theta = \beta \text{Cov}_{\theta \sim \overline \rho}\big(\overline{U}, \sigma^2_v \big) - \text{Cov}_{\theta \sim \overline \rho}\big(\overline{U}, \overline{v} - \sigma_{v,a} \big), \text{ or} \nonumber \\
    \partial_T \langle \langle \overline{U} \rangle \rangle =& \frac{2}{T^2}\text{Cov}_{\theta \sim \overline \rho}\big(\overline{U}, \overline{v} - \sigma_{v,a} \big)- \frac{4}{T^3} \text{Cov}_{\theta \sim \overline \rho}\big(\overline{U}, \sigma^2_v \big).\nonumber
\end{align}
We assume that $\overline{U}$ and $\overline{v}$ are similar enough that $\text{Cov}_{\theta \sim \overline \rho}\big(\overline{U}, \overline{v}\big)$ is positive when there is learning. If in addition $\text{Cov}_{\theta \sim \overline \rho}\big(\overline{U}, \sigma_{v,a}\big)$ is negligible we then have $\partial_\beta \langle \langle \overline{U} \rangle \rangle < 0 $ for very low $\beta$ and $\partial_\beta \langle \langle \overline{U} \rangle \rangle > 0$ for very large $\beta$ when $\text{Cov}_{\theta \sim \overline \rho}\big(\overline{U}, \sigma^2_v \big) > 0,$ as we also expect.\footnote{Taking one more derivative above yields $$
    -\partial_\beta^2 \ln \overline Z = \partial_\beta \langle \langle \overline{v} - \beta \sigma^2_v - \sigma_{v, a} \rangle \rangle = -\text{Var}_{\theta \sim \overline \rho}\big(\overline{v} - \beta \sigma^2_v - \sigma_{v, a}\big) \leq 0,$$ so increasing $T$ can only increase $\langle \langle \overline{v} - \beta \sigma^2_v - \sigma_{v, a} \rangle \rangle.$}
Combined, these considerations suggest the test performance would attain its minimum value at some intermediate value that satisfies $\partial_\beta \langle \langle \overline{U} \rangle \rangle = 0$ which can be rearranged and expressed in terms of the temperature as the implicit relation
\begin{equation}
    T_\text{opt} = 2\frac{ \text{Cov}_{\theta \sim \overline \rho}\big(\overline{U}, \sigma^2_v \big)}{\text{Cov}_{\theta \sim \overline \rho}\big(\overline{U}, \overline{v} - \sigma_{v,a} \big)} \approx 2\frac{ \text{Cov}_{\theta \sim \overline \rho}\big(\overline{U}, \sigma^2_v \big)}{\text{Cov}_{\theta \sim \overline \rho}\big(\overline{U}, \overline{v}  \big)}, \label{eq:optimalT}
\end{equation}
i.e., the optimal temperature is a trade-off that keeps $T$ large to prevent over-fitting to the variance $\sigma^2_v(\theta)$ but small enough to fit $\overline{v}.$ When the numerator and denominator in the above expression have the same sign (indeed, we expect them to be positive) and are not zero, the optimal temperature is positive and finite. We consider this a prediction in need of empirical validation, and the experiments in Figure \ref{fig:SGDMomentumCosineK32DataSplitting} are consistent with it. However, in other experimental conditions or applications we might never reach the high temperatures required to see such an effect, since there is an experimental limit above which higher temperatures will result in SGD trajectories that blow up (this limit could be found by using a batch size of 1 and the largest learning rate that prevents exploding SGD trajectories). \cite{bradley2021can} study a similar setup but without averaging over data sets and argue that loss curvature plays an important role in generalization. An analogous argument can be made here and is briefly explored in Appendix \ref{sec:testPerfAndCurvature}.

We can also study how temperature impacts the train loss, starting by noting that $ \partial_T \rho(\theta|x_\ell) =  \frac{2}{T^2}(v - \langle v \rangle)\rho(\theta|x_\ell).$ Then for any function $f(\theta, x)$ that is independent of $T$ 
\begin{align}
    \partial_T \overline{\langle f \rangle} = & \int \text{Cov}_\mathcal{D}\big(\partial_T \rho, f \big) d\theta + \partial_T \langle \langle \overline{f} \rangle \rangle
  =  \frac{2}{T^2} \overline{ \text{Cov}_{\theta \sim \rho }\big( v, f \big)}, \text{ so} \nonumber \\
    \partial_T \overline{\langle U_\ell \rangle} = & \frac{2}{T^2} \overline{ \text{Cov}_{\theta \sim \rho}\big( v, U_\ell \big)} = \frac{2}{T^2}\biggr(\text{Cov}_{\theta \sim \rho , x_\ell \sim r_\ell}\big( v, U_\ell \big) - \text{Cov}_{x_\ell}\big(\langle v \rangle, \langle U_\ell \rangle \big) \biggr).\label{eq:dTTrain}
\end{align}
We expect the right hand of Eq. \ref{eq:dTTrain} to be positive when there is learning, so increasing the temperature should only worsen the train loss, as we indeed find experimentally. 

\section{Experiments}
\label{sec:ExperimentalResults}
\subsection{Setup}
The role of SGD noise in generalization is an active research area. The diffusion approximation of SGD predicts that the noise in plain SGD can be effectively summarized by the temperature $T=\lambda/B.$  Here we empirically evaluate the main approximate predictions we obtained above:
\begin{enumerate}[label=\Alph*.]
    \item The gap decreases with $T$ and follows Eq. \ref{eq:GapVsT}.
    \item There is a finite positive optimal temperature for test performance, as suggested by Eq. \ref{eq:optimalT}.
    \item The train loss increases with temperature, as suggested by Eq. \ref{eq:dTTrain}.
\end{enumerate}
Although these were obtained by studying plain SGD, we experiment with additional SGD variants, including a learning schedule and momentum, and find that the predictions hold except when momentum is used without a learning schedule. Our experiments use a ResNet architecture on the Cifar 10 data set.  The ResNet model architecture was introduced in \cite{he2016identity}; it offers competitive results on vision classification tasks and exhibits the double-descent behavior of other large models (e.g., see \cite{nakkiran2021deep}). \footnote{We chose ResNets on Cifar 10 for our experiments rather than other architectures with similar characteristics for convenience, since this choice allowed us to use the Pytorch implementation from \cite{nakkiran2021deep} found at \url{https://gitlab.com/harvard-machine-learning/double-descent/-/blob/master/models/resnet18k.py}, itself based on the earlier implementation in \url{https://github.com/kuangliu/pytorch-cifar}. The code allows one to specify the number of convolutional filters in each ResNet block to obtain architectures with anywhere between 3k and 11M parameters, enabling future exploration of the effect of model size.} We used 32 filters in the first ResNet layer, 64 in the second and 128 in the third, resulting in a model with 2,796,202 parameters. All experiments were carried out in a Linux machine with two NVIDIA GeForce RTX 3080 GPUs. 

We consider two ways to sample data sets. (1) Data splitting: merging, reshuffling, and resplitting Cifar10 into 50,000 training images and 10,000 test images. And (2) Random sampling: uniformly sampling 50,000 training images without replacement from the 60,000 Cifar10 images, and independently similarly sampling 10,000 test images from the original 60,000 images. The latter leads to many images being shared by the training and the test set, while the former guarantees no intersection between the data sets. Random sampling reflects the assumptions in our theory with the training and test being independent, but it leads to a large variance in the results because of the high percentage of test images that is also in the training set and its variance across data set realizations. Data splitting leads to a correlation between training and test sets, e.g., if a particularly difficult example is the training set, then the test set does not have it and gets easier. So it violates the independence assumption that is key to our results, but it seems to do so mildly, and the results have a much smaller variance. We focus on data splitting experiments in the main paper, and include the analogous experiments that rely on random data set sampling in Appendix \ref{sec:moreExperiments}, along with additional experimental details.

\subsection{Results}
For each combination of data set sampling method, random seed, batch size and learning rate, we run SGD for a pre-specified number of epochs that is scaled based on a reference run that led to converging train and test losses and accuracies. We then estimate the steady-state metrics of each experiment by taking the median over the last 50 epochs of each metric, resulting in a single datapoint in our plots. The results in Fig. \ref{fig:SGDMomentumCosineK32DataSplitting} are consistent with prediction A: the gray line fits the points strikingly well, and although we do not show it, the same functional form of Eq. \ref{eq:GapVsT} fits the accuracy gap and the test loss well in these experiments. The fitted value of the parameter $c$ is much larger than 1, which means that the exponential portion of Eq. \ref{eq:GapVsT} remains at approximately $1$ and is unnecessary. Prediction B also holds, with the test loss achieving its minimum around $T=2e^{-4}$ and the test accuracy at similar value of $T.$ Prediction C is also consistent with these results. Overall, the train loss is so small for these range of temperatures that the gap and the test loss are essentially the same.  

\begin{figure}[ht]
\centering
\textbf{Plain SGD} \par
\includegraphics[width=0.95\columnwidth]{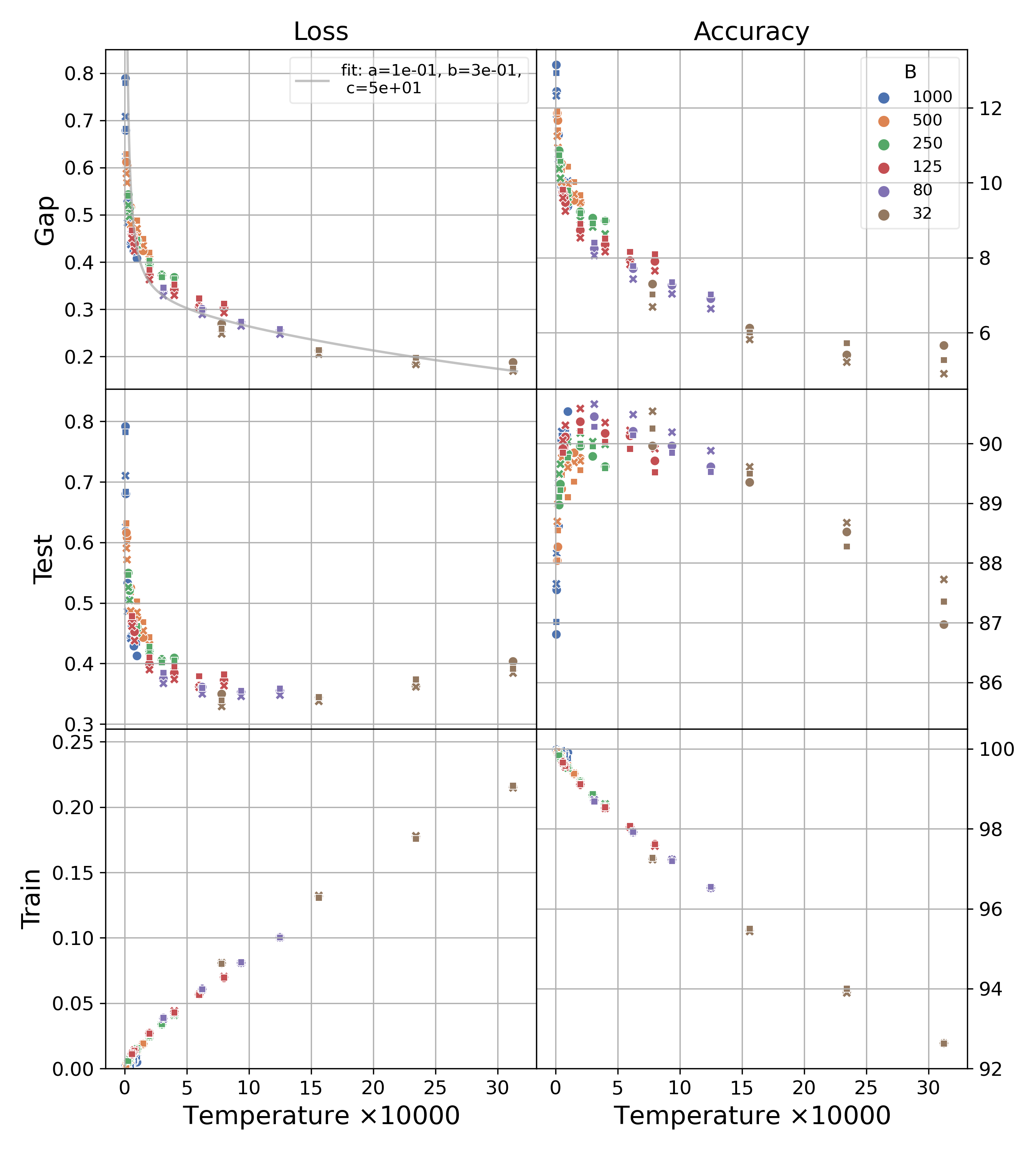}
\caption{Same setup as Fig. \ref{fig:SGDMomentumCosineK32DataSplitting} but using plain SGD: no momentum or learning schedules.}
\label{fig:PlainSGDK32DataSplitting}
\end{figure}

\begin{figure}[ht]
\centering
\textbf{Plain SGD} \par
\includegraphics[width=0.95\columnwidth]{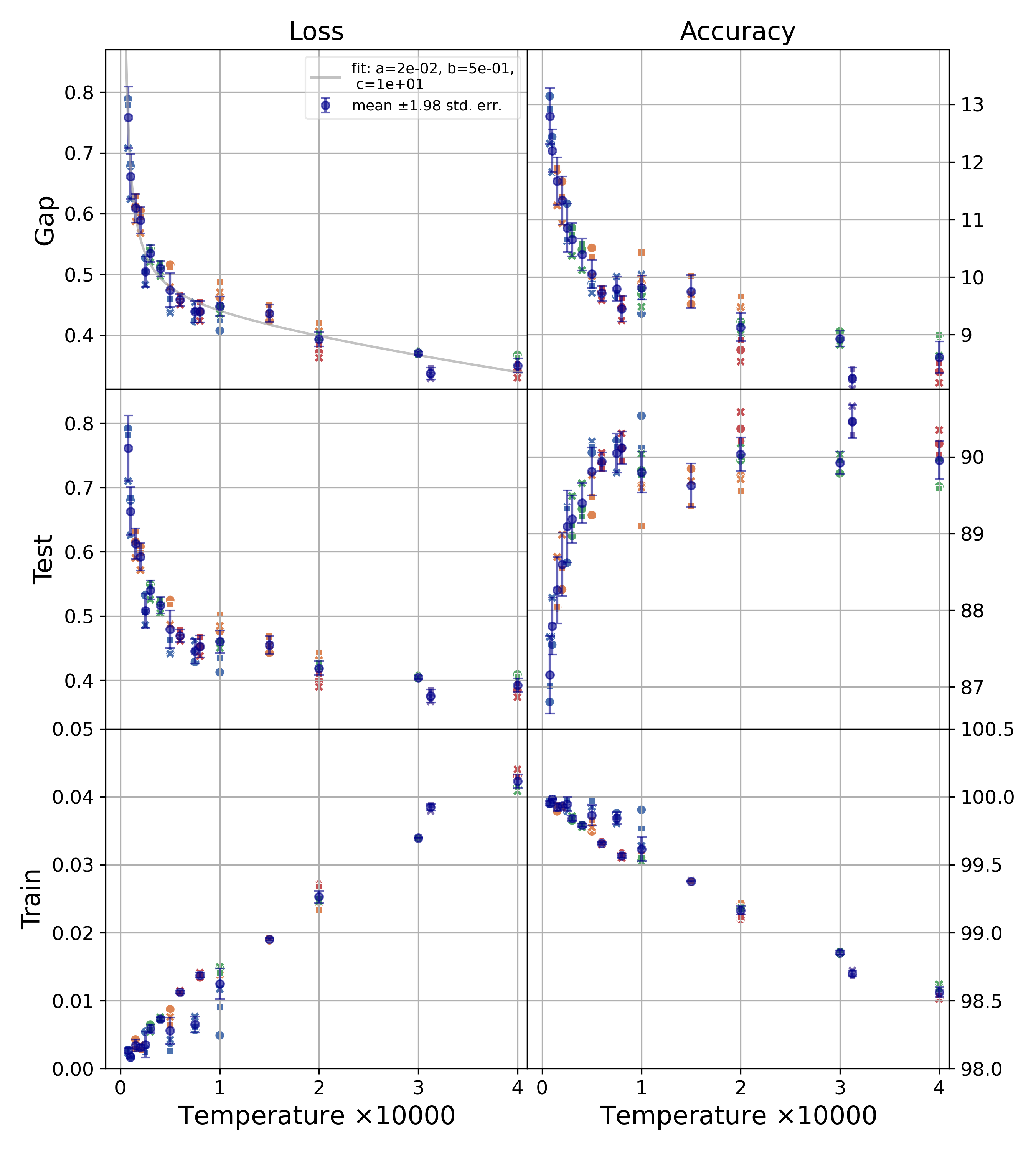}
\caption{Same experiments as Fig. \ref{fig:PlainSGDK32DataSplitting} but focused on the smaller temperatures. Eq. \ref{eq:GapVsT} was fit here only using that datapoints plotted, ignoring those for higher temperatures.}
\label{fig:PlainSGDK32DataSplittingZoomedIn}
\end{figure}

\begin{figure}[ht]
\centering
\textbf{SGD with Cosine Learning Schedule} \par
\includegraphics[width=0.95\columnwidth]{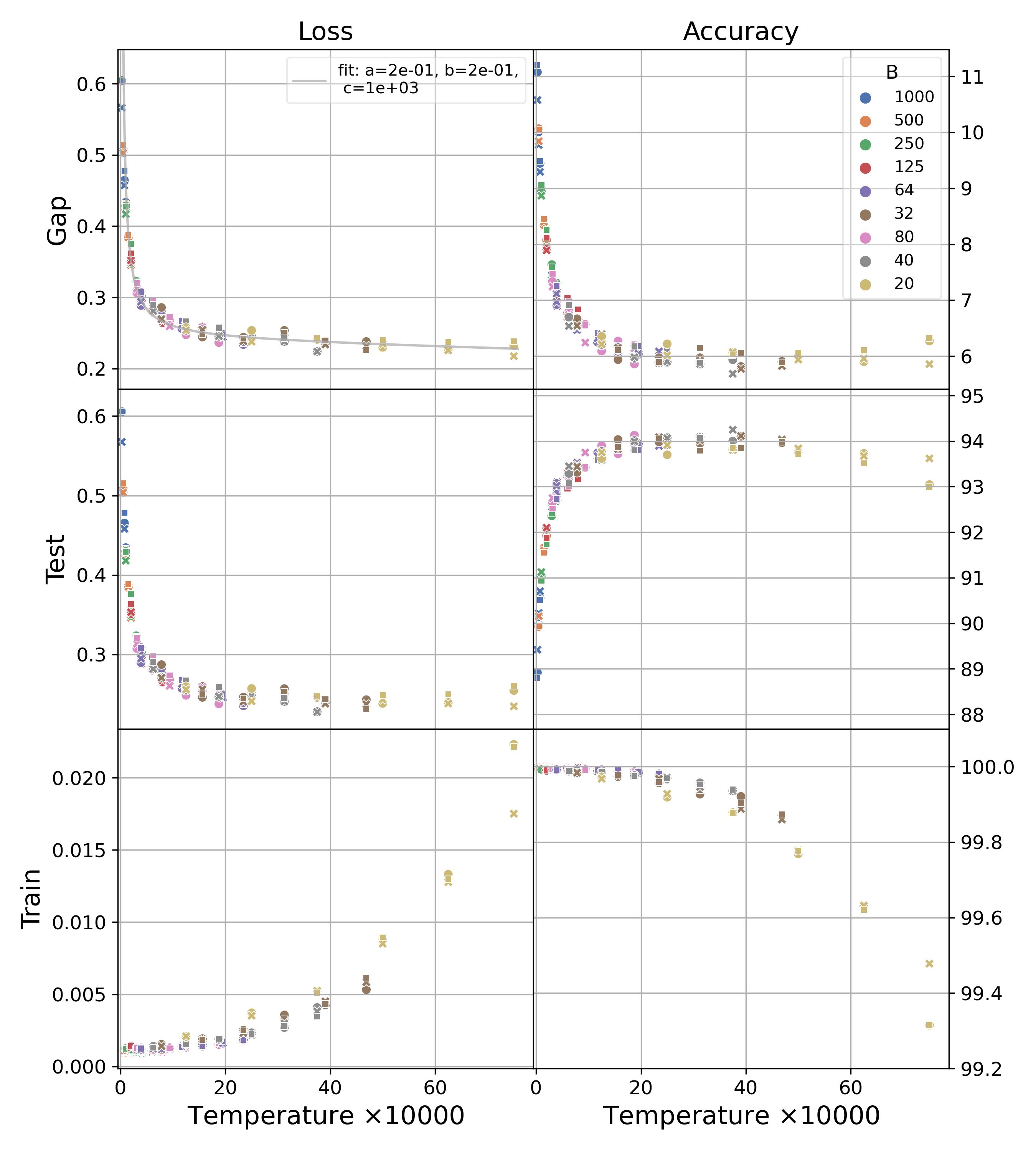}
\caption{Same setup as Fig. \ref{fig:PlainSGDK32DataSplitting} but using SGD with cosine learning schedule (no momentum).}
\label{fig:SGDCosineK32DataSplitting}
\end{figure}

\begin{figure}[ht]
\centering
\textbf{SGD with Momentum} \par
\includegraphics[width=0.95\columnwidth]{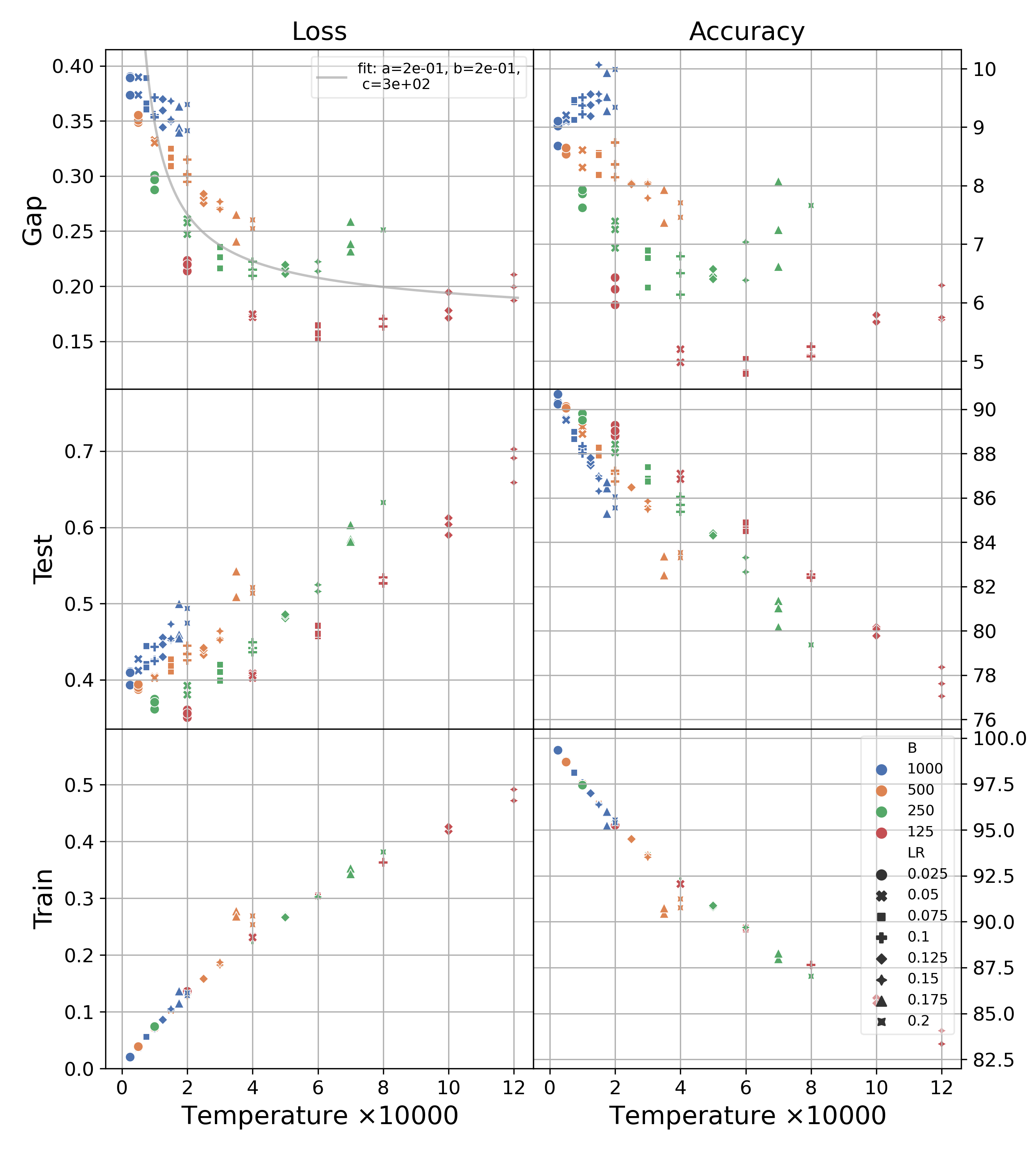}
\caption{Same setup as Fig. \ref{fig:PlainSGDK32DataSplitting} but using SGD with momentum.}
\label{fig:SGDMomentumK32DataSplitting}
\end{figure}

Figure \ref{fig:PlainSGDK32DataSplitting} shows results for plain SGD, and the three predictions appear valid here as well. But there are two differences.  At first glance the test loss and accuracy seem to have a positive optimal temperature, consistent with prediction B. But a closer inspection of Fig. \ref{fig:PlainSGDK32DataSplitting} suggests that the optimal temperature might be batch dependent, with smaller batches having a higher optimal temperature. Perhaps this is due to the validity of the SGD diffusion approximation breaking down at different values of $T$ for different batch sizes. The second main difference is that now the factor $e^{-T/c}$ in Eq. \ref{eq:GapVsT} is ``active,"  accounting for the persistent decrease in the gap as $T$ increases. Figure \ref{fig:PlainSGDK32DataSplittingZoomedIn} zooms into the lower temperature portion of the results to better appreciate the average trends as well as the variance across runs at each temperature. Figure \ref{fig:SGDCosineK32DataSplitting} shows results when using SGD with cosine learning, matching the three predictions, but we had to use batches as small as 20 to get to the higher temperatures where test performance worsens again.  Figure \ref{fig:SGDMomentumK32DataSplitting} shows results using SGD with momentum. Now experiments using different batch sizes do not line up now when plotted as a function of $T$, which can be understood through the corresponding diffusion approximation as discussed in Appendix \ref{sec:diffusionApproximations}. The evidence in favor of A is now weak at best (perhaps it holds for each batch size for low enough temperatures). Prediction B does not seem to hold; now increasing $T$ seems to worsen test performance even at very low temperatures. Prediction C is the only one that still holds; in fact the train loss and accuracy trace a rather smooth increasing function of $T$. SGD with a cosine learning schedule, with or without momentum, are the two procedures that achieve the highest test accuracy above 94\%, and the lowest test loss of about 0.25. Using only a cosine learning schedule but no momentum results in a wider range of temperatures with a close-to-optimal test performance starting with a batch size of about 80, whereas the narrower range of temperatures that achieve close-to-optimal performance when training with SGD with cosine learning and momentum needs batch sizes between 500 and 80. A batch size of 80 already significantly increases training time (per epoch, but also typically the total training time required to reach steady state) compared to using larger batch sizes. E.g., a batch size of 2000 typically took less than 2 seconds per training epoch, while a batch size of 20 took 28 seconds per epoch. So there is a trade off between test performance and training time, with better performance requiring smaller batch sizes that take longer to train. Similar observations hold for the analogous experiments ran using random sampling to construct the data sets, as shown in Appendix \ref{sec:moreExperiments}. But the variance across runs for a given $T$ is much larger. Although the fit of Eq. \ref{eq:GapVsT} to our experiments is good at low enough temperatures for most training procedures (SGD with momentum being the exception) and two data set sampling methods we consider, more studies are needed on different model architectures and data sets to see how broadly Eq. \ref{eq:GapVsT} can describe the gap. 

\subsubsection{Variance of the Loss Across Data Points}

\begin{figure}[ht]
\centering
\includegraphics[width=0.45\columnwidth]{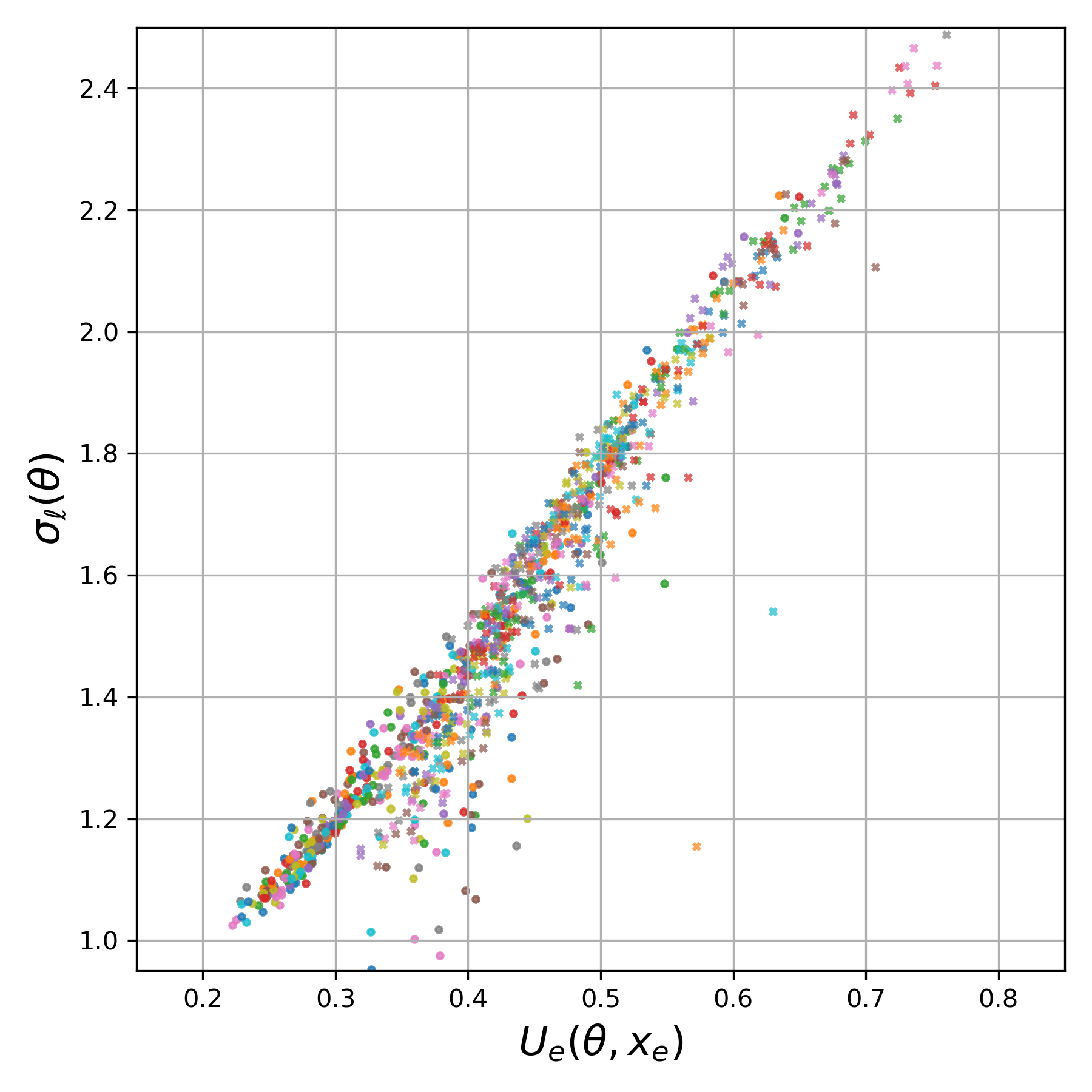}
\includegraphics[width=0.45\columnwidth]{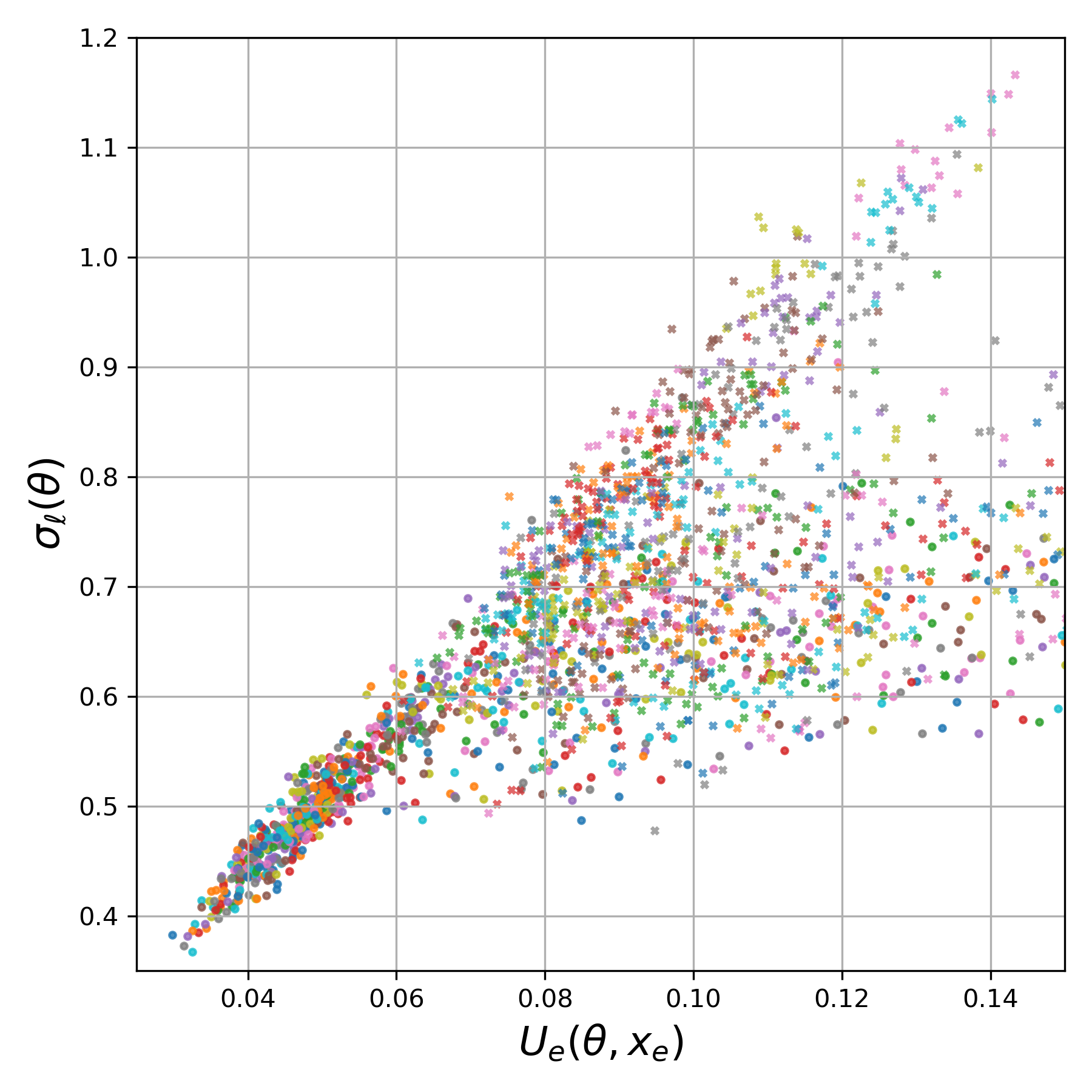}
\caption{(Left) Test set loss standard deviation versus test set loss based on the data-splitting experiments used for Figs. \ref{fig:PlainSGDK32DataSplitting} and \ref{fig:SGDMomentumCosineK32DataSplitting}. (Right) Test set loss standard deviation versus test set loss based on the random data set sampling experiments used for Figs. \ref{fig:PlainSGDK32DataAveraging} and \ref{fig:SGDCosineMomentumK32DataAveraging}. We sampled 20 evenly-separated epochs from the last $25\%$ of epochs of every experiment to plot 20 datapoints. Different colors correspond to different runs, and the two marker types to the two SGD variants used for the underlying experiments: plain SGD (circles), and SGD with momentum and cosine learning schedule (x-shaped markers).}
\label{fig:TestsetStdVsLoss}
\end{figure}

The variance of the loss $\sigma^2_\ell(\theta),$ or rather the standard deviation $\sigma_\ell(\theta),$ is one of the components in the average gap upper bound in Eqs. \ref{eq:gap_upperbnd} and \ref{eq:gap_upperbnd_approx_SGD2} and can be explored experimentally in some ways with relative ease throughout an SGD trajectory. After every training epoch, we evaluate the loss on every data point in the test set, and then compute their mean $U_e(
\theta, x_e),$ and their standard deviation. The latter is an estimate of $\sigma_\ell(\theta)$ based on the test set for the current parameter values $\theta$. Figure \ref{fig:TestsetStdVsLoss} shows scatter plots of these estimates of $\sigma_\ell(\theta)$ versus the test loss $U_e(\theta, x_e)$. For each value of the test loss, it shows several possible standard deviation values, but these seem reasonable well-concentrated around a mean value that visually appears linear in the test loss. This suggests a contribution to the gap upper bounds that is roughly linear in the test loss.  Unfortunately we do not know how to inspect $\sigma_\rho(\theta)$ or $\sigma_v(\theta),$ the other components in the gap upper bounds, for a fuller interpretation of them.

\section{Discussion}
\label{sec:Discussion}

Machine learning has been demonstrating remarkable performance in a wide range of applications including more traditional areas of computer science, like vision and natural language processing (e.g., \cite{dosovitskiy2020image, ramesh2022hierarchical, song2020score, brown2020language}), as well as fields as disparate as Biology (e.g., \cite{verma2018efficient}), Chemistry (e.g., \cite{hiranuma2021improved}), and numerical methods for physical simulation (e.g., \cite{bar2019learning, kochkov2021machine}). Typical recent successes rely on new model architectures with one or more neural network components that have as many as billions of parameters, trained on data sets that are orders of magnitude smaller than the number of model parameters, via some stochastic learning algorithm, such as stochastic gradient descent (SGD). These large models are so flexible that they often learn almost perfectly all or most of the training examples, even if these examples are all noise. Yet, the same models can still generalize well, performing very well on unseen examples when trained appropriately. Understanding how the learning algorithm, data sets, and model combine to determine generalization in such situations is an important ongoing effort, and the focus on this paper. There are at least two main research directions aimed at understanding generalization. The more common approach breaks up the analysis of test performance into two related problems that are studied separately: optimization of the training loss, and understanding the gap between test loss and training loss. These efforts have produced worst-case upper bounds on the generalization gap that are rarely predictive of typical performance when applied to practical situations, e.g., see Chapter 6 in \cite{hardt2021patterns} for an introductory overview. The second and less common direction seeks expressions for the typical train and test performance of very simple models, like a single perceptron, trained via algorithms that are not representative of those used today, and under some appropriate ``thermodynamic limit" (e.g., letting the model parameters and training set size going to infinity while keeping their ratio constant and finite); \cite{engel2001statistical, opper2001learning} are useful introductory overviews to this line of work inspired by statistical mechanics. This paper is closer in spirit to this approach in that we study typical model performance interpreted as they do, namely as the average model performance under different runs of the learning algorithm and different realizations of the data sets. We similarly rely on some ideas and techniques from statistical mechanics, only some of which are shared with previous work. The main differences with previous work include our focus on approximate results that are closer to today's applications, i.e., allowing for more complex models trained via SGD rather than through simpler algorithms that are rarely used, and understanding how the noise intrinsic to SGD impacts model performance rather than how model performance depends on training set size (although we make some remarks about this as well). The most related work is \cite{seung1992statistical} and has been discussed earlier in the paper. Generalization bounds based on the stability of SGD, such as those in \cite{hardt2016train}, also study train and test performance averaged over data set realization and model parameters. But their analysis focuses on obtaining generalization upper bounds as a function of training time and training set size; these bounds grow linearly with training time and become vacuous in typical deep learning applications, and scale inversely with data set size, which does not apply close to the interpolation regime. In contrast, our general average gap results and corresponding upper bound are not SGD specific and focus on the (co)variance of the model parameter distribution, which depends implicitly on training time and training set size, and the train loss. These properties carry over when, introducing approximations, we specialize our results to SGD at steady state, and allow us to study how SGD noise affects generalization.

We have shown that averaging over data sets can yield new useful results about some key machine learning metrics. The generalization gap is an integral that depends on the covariance between the parameters and the train loss, which can be analyzed to yield new upper bounds, and approximated in various ways, e.g., to suggest its typical dependence on the SGD noise strength that is consistent with our experiments. Also, average test depends only on the data-averaged parameter distribution $\overline{\rho}(\theta)$ and data-averaged loss $\overline{U}(\theta),$ and our approximations suggest it is optimized with SGD at a positive but finite SGD noise level, also consistent with our experiments. Similarly, average train performance only worsens with increased SGD noise. However, our results have many limitations. On the experimental side, further work can explore the behavior of generalization, test and train performance on other tasks and model architectures when training via SGD or other common algorithms for a fuller evaluation of our results. Other recent works have noted the dependence of test performance on $T$ for plain SGD, often based on a similar diffusion approximation. But our experiments are complementary. E.g., \cite{goyal2017accurate} provides ample empirical evidence using ResNets on the ImageNet data set that for a fixed $T$ and appropriate temporal rescaling of training time, different runs of plain SGD behave similarly. \cite{he2019control} use a similar diffusion approximation of plain SGD to ours under more stringent assumptions (of a locally quadratic loss and isotropic SGD noise) to derive a gap upper bound that depents on $T,$ and argue empirically that test accuracy improves as $T$ increases for ReseNets and VGG on Cifar10 and Cifar100. But they miss that there is an optimal positive temperature that optimizes test performance, do not consider any variant of SGD in their experiments, do not focus on steady state instead using a constant 200 epochs of training for all conditions, and do not consider data set sampling. \cite{bradley2021can} also shows empirical evidence of improved test performance at increased $T$ for low temperatures when using plain SGD at steady state, but do not cover the broader range of temperatures, learning algorithms or data set sampling we do here. \cite{mccandlish2018empirical} argue in an Appendix C that for a simple model with a quadratic loss and under simplifying assumptions $\lambda/B$ determines the noise scale or temperature of SGD, and state that this is consistent with their experiments.

We believe the experiments here are the first to cleanly show the dependence on $T$ of train and test metrics for a wide enough range of values to see several key effects, including the optimal finite $T$ that optimizes test performance, the robustness and limitations (SGD with momentum) of these behaviors when using SGD variants and under data set sampling, as well as the explicit functional form of the loss gap. On the theory side, our approximations only apply to plain SGD without a learning schedule or momentum as the learning algorithm, and only when run for long enough to reach steady state. Future work could seek to include any of these dimensions, as well as study other common algorithms such as ADAM. Even for plain SGD at steady state, we do not understand some of the key quantities that define the gap and test performance, including the covariance between the parameter distribution and the train loss, the variance of the losses, and essentially all aspects of the effective potential $v(\theta, x).$ In addition, the train and test accuracies and losses, as well as the accuracy gap all seem to follow smooth functions of $T,$ similar to the gap. Future theoretical work could determine or at least motivate these functions explicitly, similarly to Eq. \ref{eq:GapVsT}. Another possible research direction is the application of random matrix theory to better understand the asymptotic behavior of $\rho(\theta, x)$ and all downstream quantities, including the dependence on model size and structure and training set size. Such an approach has led to a thorough understanding of simpler models like linear regression under some situations such as full-gradient descent learning with appropriate regularization; see \cite{advani2020high, hastie2022surprises}. The approximate steady-state SGD distribution upon which many of our results rely comes from a diffusion process approximation of SGD.  Approximating learning algorithms with diffusion processes that are easier to analyze is not new: both \cite{gelfand1991recursive} and \cite{seung1992statistical} employ this strategy to study successfully some form of gradient descent with isotropic and constant-in-theta noise. Yet, while there is a better understanding now of how to obtain a diffusion process approximation to a learning algorithm, the resulting diffusion process may still be too hard to analyze and may require new techniques. E.g., for SGD with momentum or with a learning schedule their diffusion approximation is easy to obtain, e.g., see Appendix \ref{sec:diffusionApproximations}, but the author is unable to find the corresponding steady-state solutions. Many aspects of this work touch on a few concepts and techniques from statistical mechanics; there are other approaches that draw inspiration from this field to study learning that are being pursued, such as \cite{PDLT-2022} and \cite{martin2018implicit}. 

\bibliography{data_avg}

\acks{C.G.U. thanks Arwen Bradley for many useful conversations about the topics in this paper, and detailed feedback on an early draft, Josh Susskind, Preetum Nakkiran and Etai Littwin for helpful suggestions, and John Giannandrea for supporting this work.}

\newpage

\appendix

\section{Diffusion Approximations}
\label{sec:diffusionApproximations}
Consider a learning algorithm of the form
\begin{align}
    s_{t+1}=s_t + \Delta_s, \label{eq:simpleMarkov}
\end{align}
where $s_t \in R^k$ is the state, a subset of which are the model parameters $\theta_t \in R^p$. Learning algorithms work on a sepecific training set, so all quantities that derive from it are conditioned on a training set, but we leave $x_\ell$ mostly implicit here to reduce clutter. The change in state $\Delta_s$ is a random variable that typically depends on sampling and processing a subset of the training data, as well as on the current state $s_t.$ \cite{bradley2021can} shows that the above dynamics can be approximated by a diffusion process with the following Fokker-Planck and Langevin descriptions that depend on the mean $ f(s)=E[\Delta_s|s] $ and covariance $ \text{Cov}(\Delta_s|s) = C(s)$ of $\Delta_s$

\begin{align}
d s_t =& f(s)dt + G(s)dW, \text{ and} \label{eq:langevin2}\\
\partial_t \rho(s, t) 
&= - \sum_{i=1}^k \partial_{s_i} \Big\{ f_i(s) \rho(s,t)\Big\} 
 + \frac{1}{2}  \sum_{i, j=1}^k \partial^2_{s_i, s_j} \Big\{  C_{ij}(s) \rho(s,t)\Big\}\label{eq:fp2} \\
 =& - \partial_s \cdot J, \text{ where } J_i =  f_i(s) \rho(s,t) 
 - \frac{1}{2}  \sum_{j=1}^k \partial_{s_j} \Big\{  C_{ij}(s) \rho(s,t)\Big\}
\end{align} 
is the probability current vector. Here $G(s)$ is the square root of $C(s)$, i.e., $G(s)G(s)'=C(s),$ $dW$ denotes a standard Wiener noise vector of the appropriate dimension, and $J$ is the so-called probability current. The resulting diffusion process approximates $s_t$ in the sense that $\rho(s, t=t) \approx P(s_t=s)$ if both share the same initial condition. For the plain SGD in Eq. \ref{eq:SGDMod} $s=\theta,$ and letting $g= \frac{1}{B}\sum_{i}^B \partial_\theta U(x_i, \theta_t)$ be the minibatch gradient to simplify notation, we have that $\Delta_s= -\lambda \big( g + \alpha \theta_t) + \sqrt{\lambda T} \beta w_t.$ So
$f(s)=-\lambda\big( \partial_\theta  U(\theta, x_\ell) + \alpha \theta\big)$ and $C(s)= \lambda T  \big(D(\theta, x_\ell )+ \beta^2 I \big),$ which yield Eqs. \ref{eq:langevinSGD} and \ref{eq:fp} after direct substitution into the above equations. Diffusion approximations can facilitate analysis to identify effective algorithm parameters such as $T$ for plain SGD. They also provide a direct and common way to obtain semi-explicit differential equations for any quantity that depends on the learning algorithm state, which in turn provide relationships between averages of key quantities that are often derived through different methods in related literature. We show this through examples next based on the following general result. Consider any scalar quantity $\phi(s, t)$ that depends on $s$ and possibly time and consider its average $\langle \phi \rangle = \int \rho(s, t)\phi(s, t) ds$,
where we make implicit the time argument to reduce clutter. The time derivative of $\langle \phi \rangle$ can be easily found either by application of Ito's theorem, or by multiplying the FP equation above by $\phi(s, t)$ and using integration by parts followed by some algebra (e.g., see Chapter 5 of \cite{sarkka2019applied}) to yield
\begin{align}
    \frac{d}{dt}\langle \phi \rangle 
    =& \big\langle \frac{\partial \phi}{\partial t}\big\rangle +   \big\langle \partial_{s} \phi' f(s)   \big\rangle
    + \frac{1}{2} \big\langle \text{Tr}\big(C(s) \partial^2_s \phi \big)  \big\rangle. \label{eq:dynamics}
\end{align}
For plain SGD we substitute $s=\theta,$ and the appropriate $f(s)$ and $C(s)$ obtained above to get 
\begin{align}
    \frac{d}{dt}\langle \phi \rangle = & \big\langle \frac{\partial \phi}{\partial t}\big\rangle -\lambda   \big\langle \partial_{\theta} \phi' (\partial_{\theta}  U_\ell + \alpha \theta)   \big\rangle
    + \frac{\lambda T}{2} \big\langle \text{Tr}\big((D+\beta^2 I) \partial^2_{\theta} \phi \big)  \big\rangle. \label{eq:dynamicsSGD}
\end{align}

Now consider the scalar entries of $\theta,$ and $(\theta - \langle 
\theta \rangle)(\theta - \langle 
\theta \rangle)',$ where averaging the latter gives the parameter covariance $\Sigma_\theta(t)$. Applying Eq. \ref{eq:dynamicsSGD} to these choices for $\phi$ and re-organizing into a vector and matrix equation yields:
\begin{align}
    \frac{d}{dt}\langle \theta \rangle = & -\lambda \big \langle  \partial_{\theta}  U_\ell  \big \rangle - \lambda \alpha \langle \theta \rangle, \text{ and} \label{eq:dynamics_mean} \\
    \frac{d}{dt}\Sigma_\theta = & -\lambda \biggr( \big \langle  \partial_{\theta}  U_\ell  (\theta - \langle \theta \rangle)' \big \rangle + \big \langle (\theta - \langle \theta \rangle) \partial_{\theta}  U_\ell  ' \big \rangle + 2 \alpha \Sigma_\theta \biggr) + \lambda T \big \langle  D + \beta^2 I \big \rangle. \label{eq:dynamics_cov} 
\end{align}
These equations are not closed system because they generally depend on averages of other quantities. One needs to approximate them in various ways if one wants a closed system of ODEs, but we do not pursue that here. However, the equations above can already be useful, e.g., at stationarity the equation for the mean yields $\big \langle  \partial_{\theta} U_\ell + \alpha \theta \big \rangle = 0,$ and setting Eq. \ref{eq:dynamics_cov} yields a fluctuation-dissipation result that is standard in statistical mechanics (named so because it relates the dissipation driven by $D(\theta)$ to a function of the system fluctuations $\theta - \langle \theta \rangle$). Deriving these results is the main focus of \cite{yaida2018fluctuation} working with SGD directly. The covariance relation in the special case of a quadratic loss and isotropic SGD noise is part of the upper bound of the gap in \cite{he2019control}, and derived through a different analysis of a the diffusion approximation of SGD they rely on. With our framing one can easily obtain other interesting and possibly new relationships by considering other choices for $\phi(s, t).$ E.g., letting $\phi(s, t)$ be the train loss and then the test loss we find:
\begin{align}
    \frac{d}{dt}\langle U_\ell \rangle = & -\lambda \big(\big \langle | \partial_\theta  U_\ell |^2 \big \rangle + \alpha \big \langle  \theta' \partial_\theta  U_\ell \big \rangle \big) + \frac{\lambda T}{2} \text{Tr}\big(\langle  \partial_\theta^2 U_\ell  (D+\beta^2 I) \rangle \big), \text{ and} \label{eq:dynamics_train} \\
    \frac{d}{dt}\langle U_e \rangle = & -\lambda \big( \big \langle  \partial_\theta U_e' \partial_\theta U_\ell \big \rangle + \alpha \langle  \theta' \partial_\theta U_e \big \rangle \big)+ \frac{\lambda T}{2} \text{Tr}\big(\langle  \partial_\theta^2 U_e  (D+\beta^2 I) \rangle \big). \label{eq:dynamics_test}
\end{align}
Setting these equations to zero already yields new non-trivial relationships between the train and test loss gradients and Hessians and the train loss gradient covariance that hold at the SGD steady state. Analogous ODEs and steady-state relationships can be similarly obtained for other learning algorithms from the corresponding diffusion approximations.

\subsection{SGD With Momentum}
We now work trhough the diffision approximation of the SGD with momentum in Eq. \ref{eq:SGDMomentum} because it helps explain why the results in Figs. \ref{fig:SGDMomentumK32DataSplitting} and \ref{fig:SGDMomentumK32DataAveraging} do not line up for different batch sizes as a function of $T$ any more. The state $s_t$ consists of $\theta_t$ and $v_t,$ which we can rewrite in the standard form of Eq. \ref{eq:simpleMarkov} as follows: 
\begin{align}
    \theta_{t+1} = & \theta_t  -\lambda \big(\mu v_t + g +\alpha \theta_t \big) \\
    v_{t+1}=& v_t - (1-\mu)v_t + g  + \alpha \theta_t. \label{eq:dynSGDMomentum}
\end{align}
So
\begin{align}
    f(s) = &
    \begin{bmatrix}
        -\lambda \big( \mu v_t + \partial_\theta U_\ell(\theta_t, x_\ell) + \alpha \theta_t\big)  \\
        - (1-\mu)v_t +\partial_\theta  U_\ell(\theta, x_\ell) + \alpha \theta_t
    \end{bmatrix}, \text{ and}  \\
   C(s) =&\begin{bmatrix}
        \lambda & 1 \\
        1 & \frac{1}{\lambda}
    \end{bmatrix} \otimes  T  D(\theta),
\end{align}
where $ D(\theta)$ and $T$
 are the same as in vanilla SGD, and $\otimes$ denotes the Kronecker product of matrices, making the above covariance block diagonal with four blocks proportional to $ T  D(\theta)$.
 The square root of the covariance is then
 \begin{align}
     G(s) = &  = \begin{bmatrix}
        1 \\
        \frac{1}{\lambda}
    \end{bmatrix} \otimes \sqrt{ \lambda T}  D(\theta)^{\half}.
 \end{align}
 Substitution into Eq. \ref{eq:langevin2} yields the Langevin description
 \begin{align}
     d\theta = &  -\lambda \big( \mu v_t + \partial_\theta U_\ell(\theta_t, x_\ell) + \alpha \theta_t\big) dt + \sqrt{\lambda T}  D(\theta)^{\half}dW, \nonumber \\
     dv =& \big((\mu-1)v_t +\partial_\theta  U_\ell(\theta, x_\ell) + \alpha \theta_t \big) dt + \frac{1}{\sqrt{B}}  D(\theta)^{\half}dW. \label{eq:LangevinSGDMomentum} 
 \end{align}
 Note that both $\theta$ and $v$ share the same noise source $dW \in R^n$ because of the structure of the covariance matrix of $s.$ We can re-scale time via $t'=\lambda t$ in the same way we did for plain SGD to remove $\lambda$ from the equation for $\theta$ above because both the drift term (that multiplies $dt$ in Eq. \ref{eq:LangevinSGDMomentum}) and the diffusion term (the square of the term that multiplies $dW$) are proportional to $\lambda$; but this does not work for the Langevin equation for $dv$: that one has a drift term independent of $\lambda$ and a diffusion term that depends on  $B$. This suggests that different batch sizes will lead to different traces as a function of $T,$ as we see in our experiments of SGD with momentum in Figures \ref{fig:SGDMomentumK32DataSplitting} and \ref{fig:SGDMomentumK32DataAveraging}. In some more detail, letting $t'=\lambda t$ transforms the Langevin above into
  \begin{align}
     d\theta = &  -\big( \mu v_t + \partial_\theta U_\ell(\theta_t, x_\ell) + \alpha \theta_t\big) dt' + \sqrt{T}  D(\theta)^{\half}dW', \nonumber \\
     dv =& \frac{1}{\lambda}\big((\mu-1)v_t +\partial_\theta  U_\ell(\theta, x_\ell) + \alpha \theta_t \big) dt' + \frac{1}{\sqrt{B \lambda}}  D(\theta)^{\half}dW'. \label{eq:LangevinSGDMomentum2} 
 \end{align}
 Since this is the only transformation of time that removes the dependence of the learning rate from the first equation, but then creates one in the second equation that affects the dynamics while leaving the $B$ dependence of the steady state untouched because of the diffusion term in the momentum equation, we conclude there is no transformation that can remove the learning rate and mini-batch dependencies from both equations simultaneously. This means that the system does not simply depend on $T,$ but rather on $B$ and $\lambda$ separately. One can also at this point write down the Fokker-Planck equation and reach a similar conclusion, but we omit that for brevity. The version of SGD with momentum reflects the Pytorch implementation, which is what we use for our experiments. But it is somewhat different from the more typical one in \cite{sutskever2013importance} described by
 \begin{align}
    v_{t+1}=\mu v_t - \lambda g + -\lambda \alpha \theta_t & \text{, and } \theta_{t+1} = \theta_t + v_{t+1} . \label{eq:SGDMomentumClassic}
\end{align}
 Going through an analogous argument to obtain the diffusion approximation and rescaling time with the same transform results in the following Langevin description:
  \begin{align}
     d\theta = &  \big( \frac{\mu}{\lambda} v_t - \partial_\theta U_\ell(\theta_t, x_\ell) - \alpha \theta_t \big) dt' + \sqrt{T}  D(\theta)^{\half}dW', \nonumber \\
     dv =& \big(\frac{1}{\lambda}(\mu-1)v_t -\partial_\theta  U_\ell(\theta, x_\ell) - \alpha \theta_t \big) dt' + \sqrt{T}  D(\theta)^{\half}dW'. \label{eq:LangevinSGDMomentum3} 
 \end{align}
 Even though the diffusion term only depends on $T$ as in plain SGD, the drift term is different for different learning rates even if $T$ is the same (including at steady state), so we would again not expect experiment metrics to match for different learning rates and minibatch combinations when plotted as a function of $T.$ Rather, we would expect that experiments for different minibatch sizes that hold the learning rate constant would line up well as a function of $T$; future work could evaluate this. Lastly, we note that we have not been able to find the steady-state solution either of the diffusion processes above that describe SGD with momentum, and we do not know why adding a cosine learning schedule on top of momentum makes our experiments again line up nicely as a function of $T.$
 
\section{The Steady-State SGD Distribution}
\label{sec:SGDSS}
Here we simply check that the Boltzmann-like distribution with the effective potential in Eq. \ref{eq:SGDeffPot} is an approximate steady-state SGD parameter distribution. The SGD algorithm in Eq. \ref{eq:SGDMod} can be shown to be well approximated under low enough temperatures by a diffusion process (e.g., see \cite{li2017stochastic, bradley2021can}) governed by the following Fokker-Planck equation that is equivalent to the Langevin description in Eq. \ref{eq:langevinSGD}:
\begin{align}
\partial_t \rho(\theta, t| x_\ell)
&= \lambda \biggr( \sum_{i=1}^p \partial_i \Big\{\rho(\theta,t| x_\ell) (\partial_i  U_\ell + \alpha \theta_i) \Big\}  + \frac{1}{2} T \sum_{i, j=1}^p \partial^2_{ij} \Big\{ \big(D_{ij}(\theta) + \beta^2 \delta_{ij} \big) \rho(\theta,t| \mathcal{D})\Big\}\biggr) \nonumber \\
& =\lambda \partial_\theta \cdot \bigr\{  \rho (\partial_\theta U_\ell + \alpha \theta)  +  \frac{T}{2}\rho \big(\partial_\theta \cdot D \big) +  \frac{T}{2}(D+\beta^2 I) \partial_\theta \rho \bigr\} = -\lambda \partial_\theta \cdot J. \label{eq:fp}
\end{align}
The continuous time $t$ here corresponds to the number of SGD updates, and can be redefined to ``absorb" the learning rate, i.e., letting $t'=t \lambda$ has the effect of leaving $T$ and $\alpha$ as the only parameters when re-writing the equation above in terms of $t'$ instead of $t$. So two different SGD runs with the same $T$ and $\alpha$ approximately produce the same parameter distribution as long as $n_1 \lambda_1 = n_2 \lambda_2,$ where $n_1$ and $n_2$ are the number of SGD updates in the two runs, and $\lambda_1$ and $\lambda_2$ are the learning rates used. Equivalently, but also useful, we can multiply this relation by $B_1/B_1$ on the left and $B_2/B_2$ on the right to obtain $T_1(B_1n_1) = T_2 (B_2 n_2),$ implying two runs at different temperatures are approximately comparable if the number of training examples processed ($nB$) times the $T$ is equal for the two runs. We use this to guide the number of epochs we run SGD for in our experiments. Once we find a number of epochs that yields train and test losses and accuracies that reach steady state, we scale the number of epochs of other runs (up to a point to prevent runs that take too long) by the ratio of thier temperature relative to the reference temperature. Specifically, we determine the number of epochs $N$ for each of our runs via
\begin{equation*}
    N = 300\max \big(1, \min \big(4, 1 + \frac{1}{4}(2e^{-4}/T - 1) \big) \big).
\end{equation*}
Here 300 is the reference number of epochs that produced runs with convergent metrics at the reference temperature of $2e^{-4}.$ Above, $\theta_i$ denotes the $i$-th entry of $\theta,$ $\partial_i$ the partial derivative with respect to $\theta_i$, and $\delta_{ij}$ is the Kronecker delta. 
When $\rho$ is of the Boltzmann form specified by  Eq. \ref{eq:SGDeffPot}, direct calculations yield
$\partial_\theta \rho = -\rho \big(\frac{2}{T} \partial_\theta v + \partial_\theta a \big).$
Substituting this into Eq. \ref{eq:fp} we find that
\begin{align}
    J =-\rho \biggr( \big(\partial_\theta U_\ell +\alpha \theta - (D+\beta^2I) \partial_\theta v \big) + \frac{T}{2}\big[\big(\partial_\theta \cdot D \big) - (D+\beta^2I) \partial_\theta a \big]\biggr).
\end{align}
A distribution for which $J=0$ for all $\theta$ is a steady-state solution of the diffusion process above, since then $\partial_t \rho(\theta, t| x_\ell)=0$. From the previous equation, this is the case if we set
\begin{align}
    (D+\beta^2I) \partial_\theta v =& \partial_\theta U_\ell + \alpha \theta, \text{  and }
    (D+\beta^2I) \partial_\theta a = \big(\partial_\theta \cdot D \big), \label{eq:implicit_v_a}
\end{align}
which is exactly how our effective potential is defined; see Eqs. \ref{eq:v} and \ref{eq:a}. For the line integrals in these equations to be path independent we also need their integrands to be the gradients of some scalar function. Lastly, we assume the diffusion is ergodic so that the steady state distribution is unique. The same method to obtain diffusion process approximations can be applied to other algorithms, such as SGD with momentum, or Adam. But the resulting diffusion process may still be too hard to analyze, e.g., we have been unable to obtain the steady state distribution for the diffusion approximation of SGD with momentum.

\section{Delta Method Approximation Of The Gap}
\label{sec:appDeltaMethod}

The delta method is a standard technique in statistics that approximates the average of non-linear function of a random variable via a second-order Taylor series about the random-variable's mean before the expectation is taken. Denote the random variable by $x$, and its mean and covariance by $\mu$ and $\Sigma.$ Applying the second-order delta method to the covariance of two functions, $f(\theta, x)$ and $h(\theta, x),$ results in:

\begin{align}
    \text{Cov}_x \big(f, h \big) \approx & \half \biggr(\text{Tr}\big(\Sigma\big[\partial_x f \partial_x h^t + \partial_x h \partial_x f^t  \big]\big) - \half  \text{Tr}\big(\Sigma \partial^2_x f \big)\text{Tr}\big(\Sigma \partial^2_x h \big) \biggr), \label{eq:cov1}
\end{align}
where the derivatives are evaluated at the mean $\mu.$ We can directly evaluate the above when $f=\frac{1}{Z}e^{-g(\theta, \mathcal{D})}=\rho,$ and $g=U_\ell$ by letting $x=\{ g(\theta, x_\ell), U_\ell(\theta, x_\ell)\}$ to obtain
\begin{equation}
    \text{Cov}_\mathcal{D}(\rho, U_\ell)  \approx -\rho(\theta|\mu)\sigma_{g, u}(\theta),
\end{equation}
where $\sigma_{g, u}(\theta)=\text{Cov}_{x_\ell}(g, U_\ell)$ is the covariance across trainset realizations of the effective potential and the train loss, and $\rho(\theta|\mu)$ is the conditional parameter distribution evaluated at the mean of the training set. We also ignored the stochasticity in $Z$ assuming it is negligible. Substituting in Eq. \ref{eq:gap_bar_brak2} then yields
\begin{equation}
    \overline{\langle \Delta \rangle} = \int \rho(\theta|\mu) \sigma_{g,u}(\theta) d\theta,\label{eq:gapApproxDelta}
\end{equation}
which is the expected covariance between $g$ and $U_\ell$ under $\rho(\theta|\mu).$ One would expect that successful approaches to learning would yield $\sigma_{g, u}(\theta) \geq 0,$ in which case the approximation of the gap is non-negative. For the SGD steady-state distribution the above becomes
\begin{equation}
    \overline{\langle \Delta \rangle} = \frac{2}{T}\langle\sigma_{v,u} \rangle_{\rho(\theta|\mu)} + \langle\sigma_{a,u} \rangle_{\rho(\theta|\mu)},\label{eq:gapApproxDeltaSGD}
\end{equation}
where $\langle f \rangle_{\rho(\theta|\mu)}$ denotes the average of any function $f(\theta)$ under $\rho(\theta|\mu).$

\section{Test Performance And Curvature}
\label{sec:testPerfAndCurvature}
\cite{bradley2021can} argues that curvature of the loss impacts test performance by analyzing the test loss averaged over $\rho(\theta|x_\ell)$ (i.e., we did not average over $x_\ell$ then). Here we briefly describe how the same intuition carries over to the total average scenario, although now the curvature is that of the data-averaged losses and potentials.  We now consider a situation where $\bar \rho (\theta)$ is a mixture of Gaussians with one mixture component for each local minima of $\bar U_\ell,$ and where each local minima of $\bar U_\ell$ has a nearby local minima of $\bar U_e,$ and where all local minima have positive curvature, i.e., are locally quadratic. This strong set of assumptions is made to get some intuition, even though we know they may not hold in some practical situations, e.g., local minima tend to not be quadratic for overparametrized models. Moving along, we index local minima with a $k$ subscript, and denote by $\tilde \theta_k$ the local minima of $\bar U_\ell(\theta),$ by $\theta_k$ the local minima of $\bar U_e(\theta),$ and assume $\bar \rho(\theta) = \sum_k w_k \mathcal{N}\big(\mu_k, \Sigma_k \big).$ We call $b_k=\mu_k - \tilde \theta_k$ the local bias, and $s_k = \tilde \theta_k - \theta_k$ be the local (train/test) shift due to distribution shift, so that $\theta - \theta_k = \theta - \mu_k + b_k + s_k.$ We also let $U_k=\bar U_e(\theta_k),$ and $C_k = \partial_\theta^2 \bar U_e(\theta_k)$ be the local test loss and its curvature, and similarly for the train loss, we let $\tilde U_k=\bar U_\ell(\tilde \theta_k),$ and $\tilde C_k = \partial_\theta^2 \bar U_\ell(\tilde \theta_k).$ Taylor expanding $\bar U_e$ and $\bar U_\ell$ around each local minima and taking expectation then yields after some algebra:
\begin{align}
    \langle \langle \bar U_e \rangle \rangle \approx & \sum_k w_k \biggr(U_k + \frac{1}{2}\text{Tr}\big(\Sigma_k C_k \big) + \frac{1}{2}(b_k + s_k)'C_k(b_k + s_k)\biggr), \\
    \langle \langle \bar U_\ell \rangle \rangle \approx & \sum_k w_k \biggr(\tilde U_k + \frac{1}{2}\text{Tr}\big(\Sigma_k \tilde C_k \big) + \frac{1}{2}b_k ' \tilde C_kb_k \biggr).
\end{align}
So the test loss is a trade-off between local minima depths and local minima curvatures, where the latter get picked up by the local variance, as well as by the bias and the train/test shift. Note this shift is between $\bar U_\ell$ and $\bar U_e$, i.e., due to a true underlying distribution shift and not sampling of the data sets. Yet sampling train/test shift also pays a curvature penalty. E.g., let $U(\theta, x) = h(\theta - s(\theta, x)) + \eta(\theta, x),$ where $s(\theta, x)$ is sampling shift, and $\eta(\theta, x)$ is just noise. Assume $\overline s(\theta) = \overline \eta(\theta)=0.$ Taylor expanding about $s=0$ and averaging, we find that
\begin{align*}
    \overline U(\theta) \approx h(\theta) + \frac{1}{2}\text{Tr}\big(\partial_\theta^2 h (\theta) \text{Cov}_x(s) \big).
\end{align*}
So curvature hurts on all fronts: sampling shift, distribution shift, bias, and parameter variance. We want models and methods that result in low curvature solutions.

\section{More Experimental Details And Results}
\label{sec:moreExperiments}
\begin{figure}[ht]
\centering
\includegraphics[width=0.45\columnwidth]{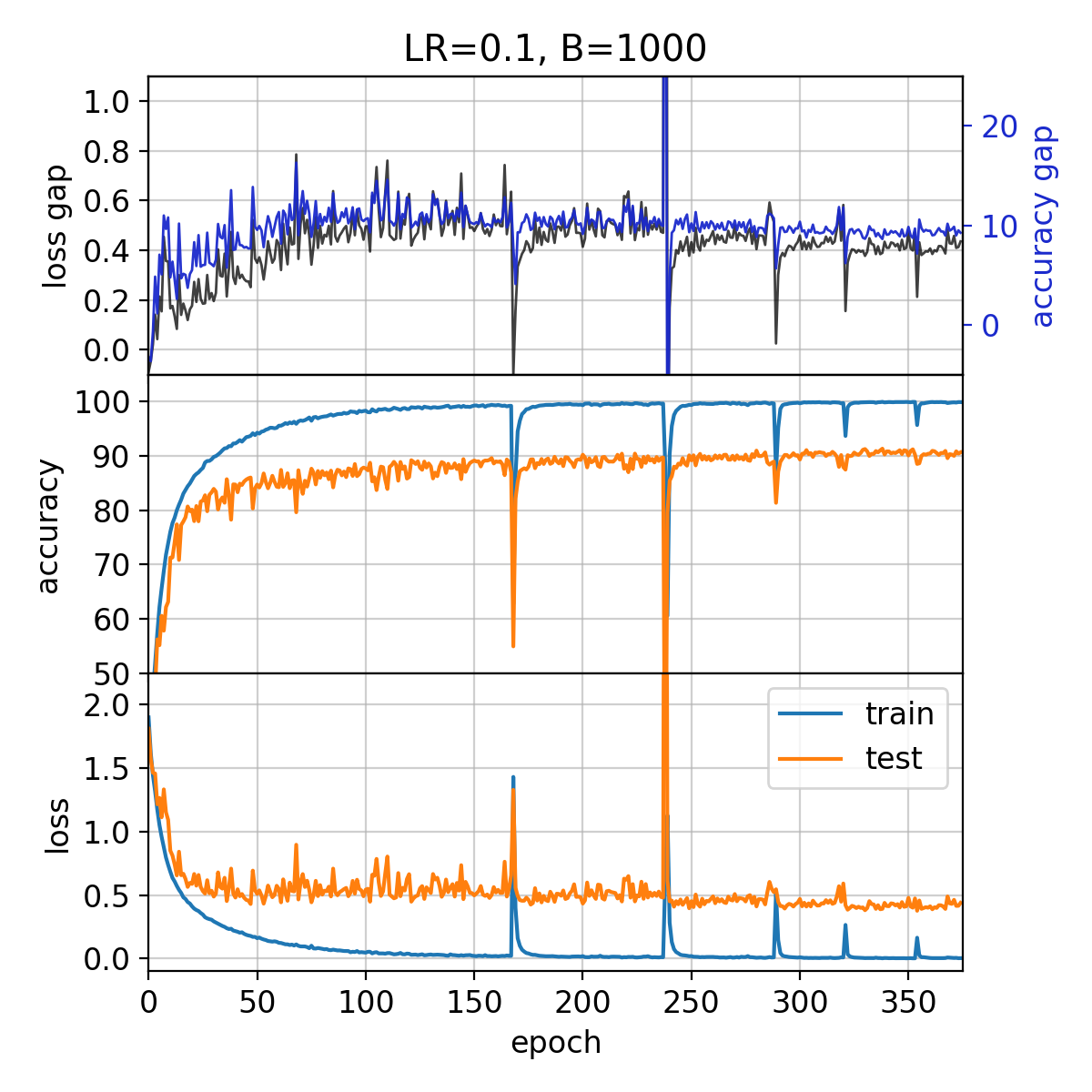}
\includegraphics[width=0.45\columnwidth]{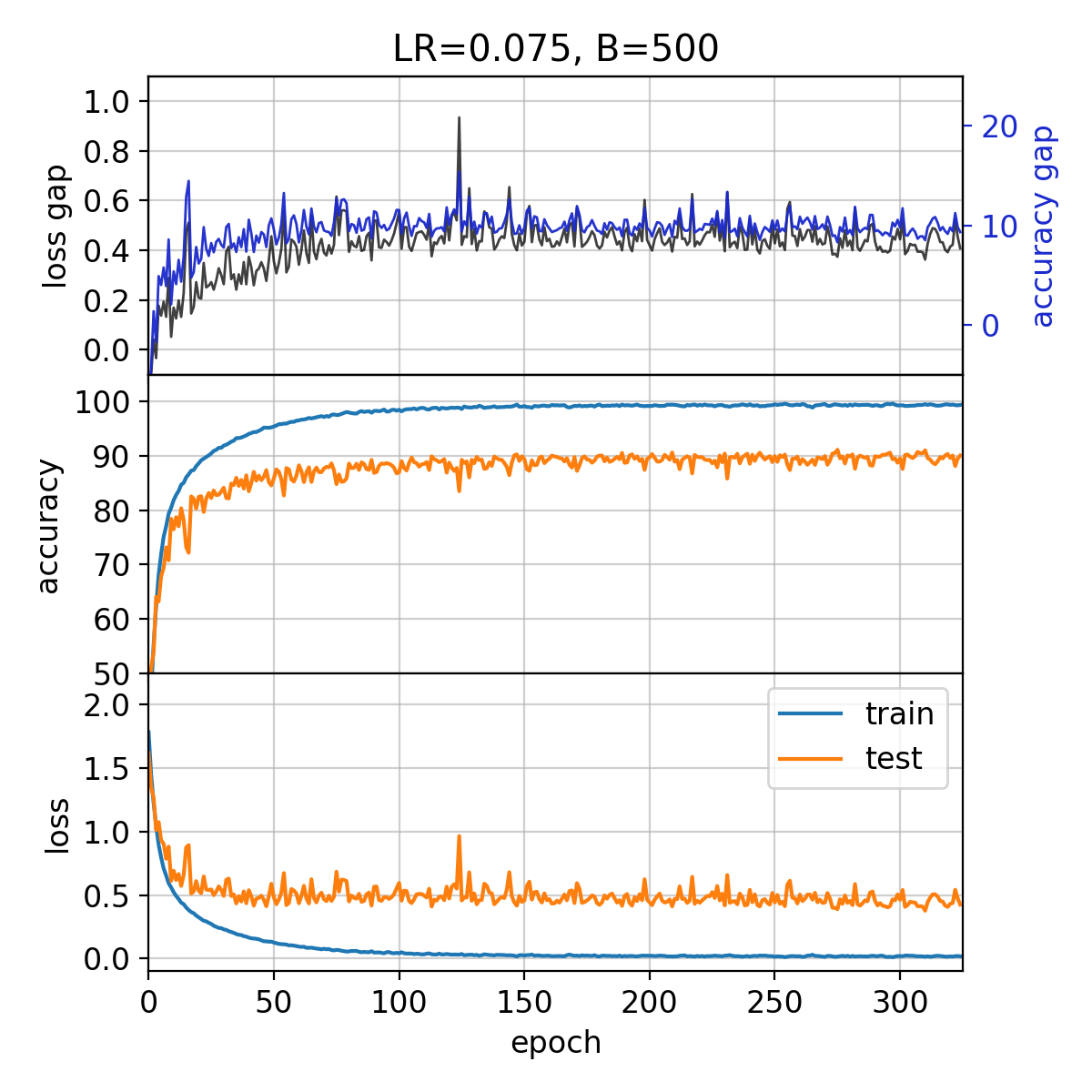}
\caption{(Left) Plain SGD under data splitting showing unstable dynamics in the form of spikes. These seem to arise only for large batch size and large learning rate combinations. (Right) Plain SGD run that is better behaved.}
\label{fig:RunWithSpikes}
\end{figure}

The diffusion approximation of SGD allows us to determine the number of epochs for different runs at different temperatures so that they are comparable in terms of "effective training time" (epochs should scale inversely with $T$; see Appendix \ref{sec:SGDSS}). We use this to scale the number of epochs in our experiments for different choices of $B$ and $\lambda$ to reach SGD steady state while avoiding running experiments for too many epochs.
The range of temperatures we use is determined by the following constraints: quality of the diffusion approximation, and number of epochs required for convergence to steady state. The former breaks down for large enough learning rates which lead to losses that become so large they result in numerical overflow. So we used learning rates of $0.1$ and as small as $0.0075.$ Also, interestingly, for large batch sizes greater than 1000 and large enough learning rates spikes in the dynamics appear that make it hard to extract accurate steady-state statistics, and for large enough learning rates. We assess convergence to steady state approximately by checking that the train loss and accuracy, and the test loss and accuracy, do not change significantly over a few tens of epochs. Having these metrics not change over epochs is clearly just a necessary condition for steady-state, but is not sufficient. However, it is enough for our purposes. Once we find a large enough number of epochs for a given learning rate and batch size that achieves this but that is otherwise a small as possible to minimize compute time, we use this as a reference number of epochs and a reference temperature for other runs at different temperatures as explained earlier. The result seems to work in the sense of obtaining converging SGD dynamics. Failure to converge to a steady state leads to all kinds of artifacts (such as gaps that get worse as $T$ increases from close to zero), and needs to handled with care. A typical reference number of epochs for our experiments at a temperature of $2e^{-4}$ was 300 epochs. Each training epoch took under one second for batch sizes of 1000 or larger, and 30 seconds for a batch size of 32, so we avoided low batch sizes as much as possible. We also note empirically that smaller models (not shown) require more epochs for convergence, as also observed elsewhere, e.g., in \cite{li2020train}. Our experiments do not change other details in the Pytorch implementation we used. We kept the use of batch norm, and data augmentation based on cropping and horizontal flips for both training and test (rather than just for train, since in our theory we want training and test to be sampled from the same underlying distribution). When extracting statistics from a single SGD run, we chose the median of the last 50 epochs rather than the mean because it is more robust against outliers, which can be extreme for larger batch sizes and learning rates that lead to large spikes. However, using the mean instead of the median did not change the results much, perhaps because we also attempt to find spikes towards the end of a run for batch sizes of 1,000 or larger, and drop the last part of the run after the spike starts. We detect spikes simply by checking if the last 7 epochs in the run have a larger average than the last 25, and detect the epoch where this starts to happen to make that the new end of the run. Figure \ref{fig:RunWithSpikes} shows an example of a plain SGD run with the unstable spikes we aim to remove before calculating our steady-state metrics, as well as a well-behaved run. We use the curve fit function in scipy.optimize to fit Eq. \ref{eq:GapVsT} to the generalization gap in our plots via minimization of squared errors.

\begin{figure}[ht]
\centering
\textbf{Plain SGD} \par
\includegraphics[width=0.95\columnwidth]{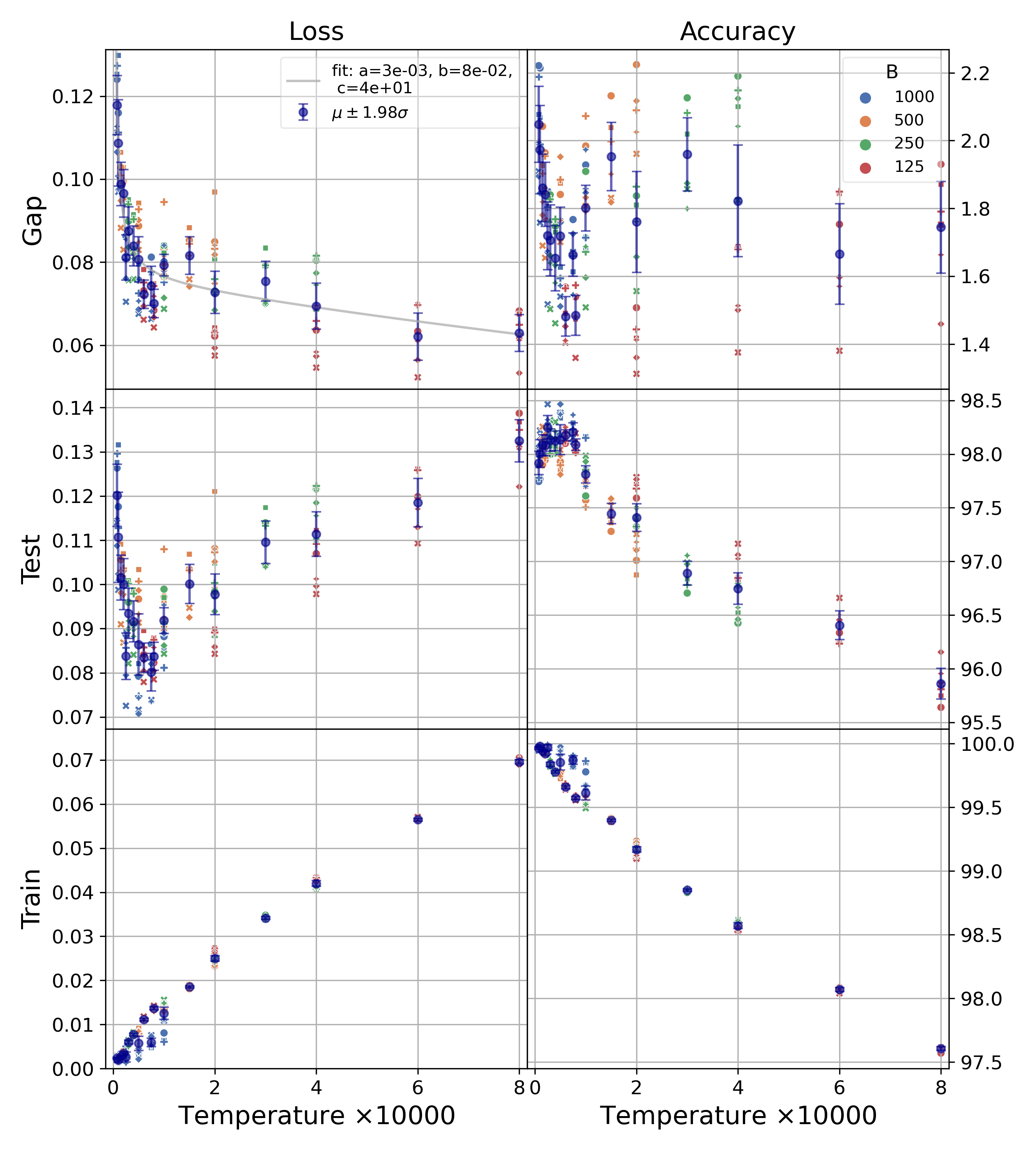}
\caption{Same setup as Fig. \ref{fig:PlainSGDK32DataSplitting} except using random sampling to generate the data sets for each experiment: the train and test sets were sampled independently using random sampling without replacement from the 60,000 Cifar10 examples. The variance across runs is very high compared to the sampling by random splitting because of the high and variable number of test examples that are also in the training. But the mean follows the expected trends: gap decreases with $T,$ there is an optimal finite positive $T$ that minimizes the test loss, and the train loss only increases with $T.$}
\label{fig:PlainSGDK32DataAveraging}
\end{figure}

\begin{figure}[ht]
\centering
\textbf{SGD with Momentum and Cosine Learning Schedule} \par
\includegraphics[width=0.95\columnwidth]{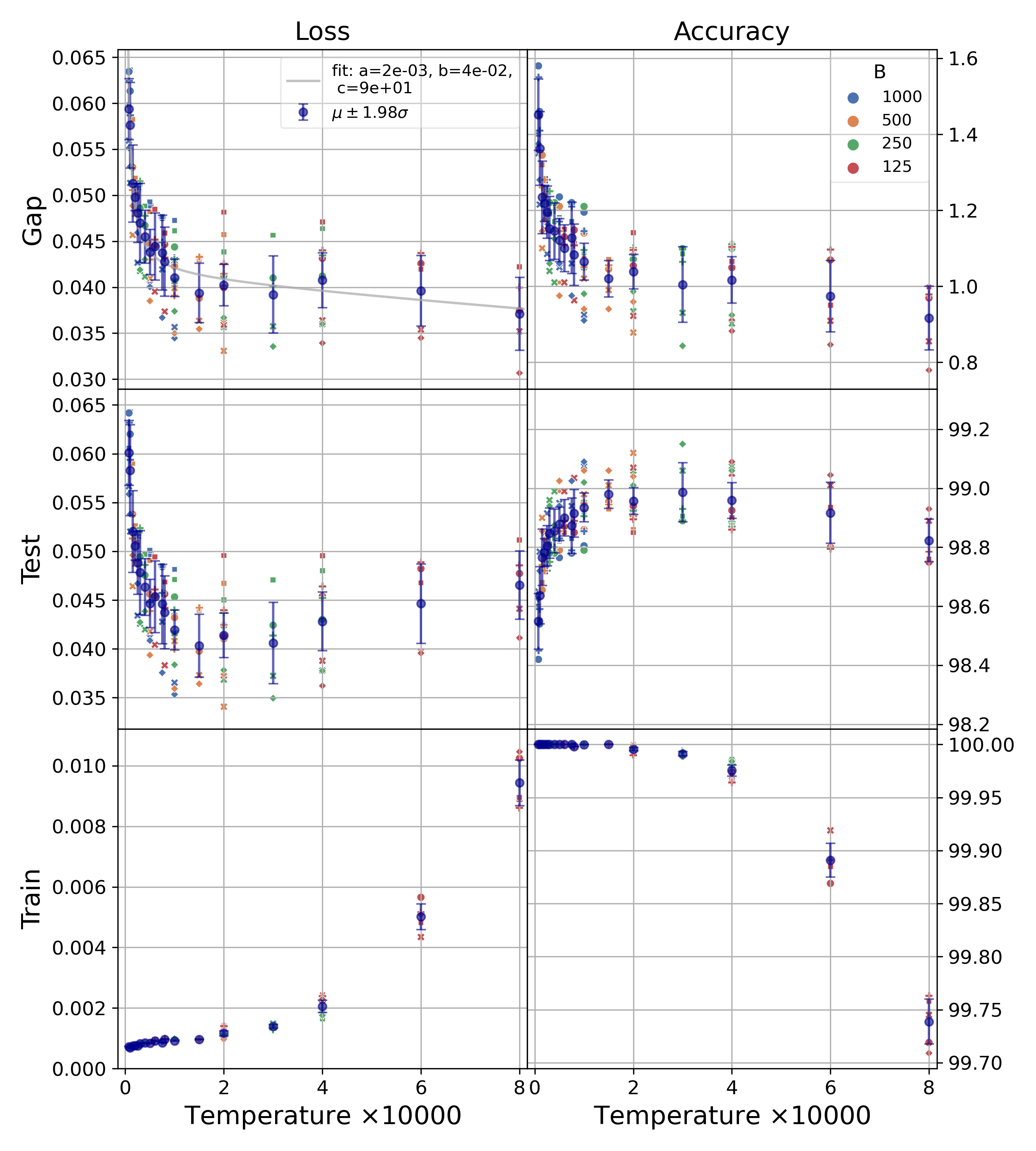}
\caption{Same setup as Fig. \ref{fig:PlainSGDK32DataAveraging} but using SGD with momentum and cosine learning schedule.}
\label{fig:SGDCosineMomentumK32DataAveraging}
\end{figure}

\begin{figure}[ht]
\centering
\textbf{SGD with Cosine Learning Schedule} \par
\includegraphics[width=0.95\columnwidth]{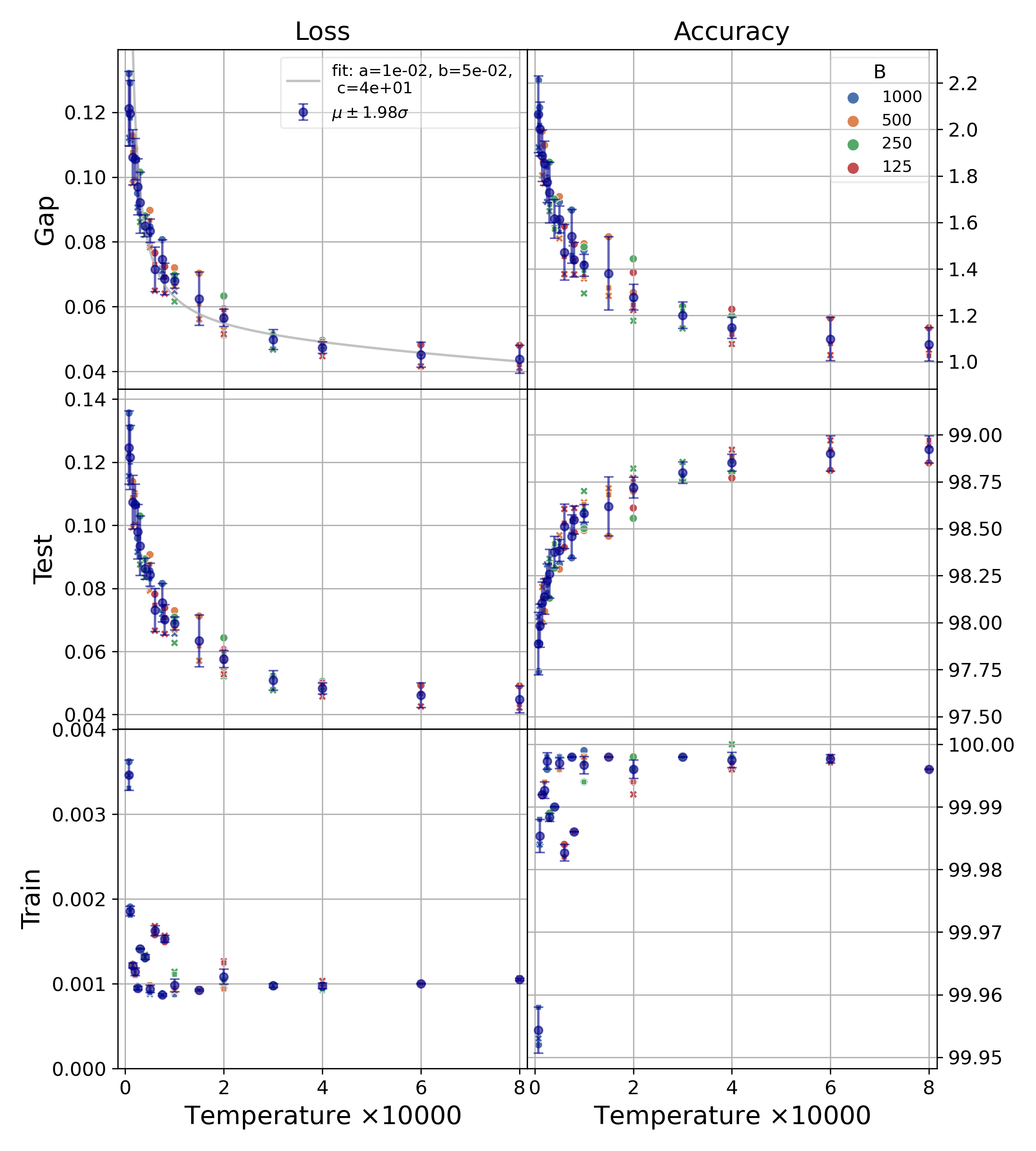}
\caption{Same setup as Fig. \ref{fig:PlainSGDK32DataAveraging} but using SGD with cosine learning schedule (no momentum).}
\label{fig:SGDCosineK32DataAveraging}
\end{figure}

\begin{figure}[ht]
\centering
\textbf{SGD with Momentum} \par
\includegraphics[width=0.95\columnwidth]{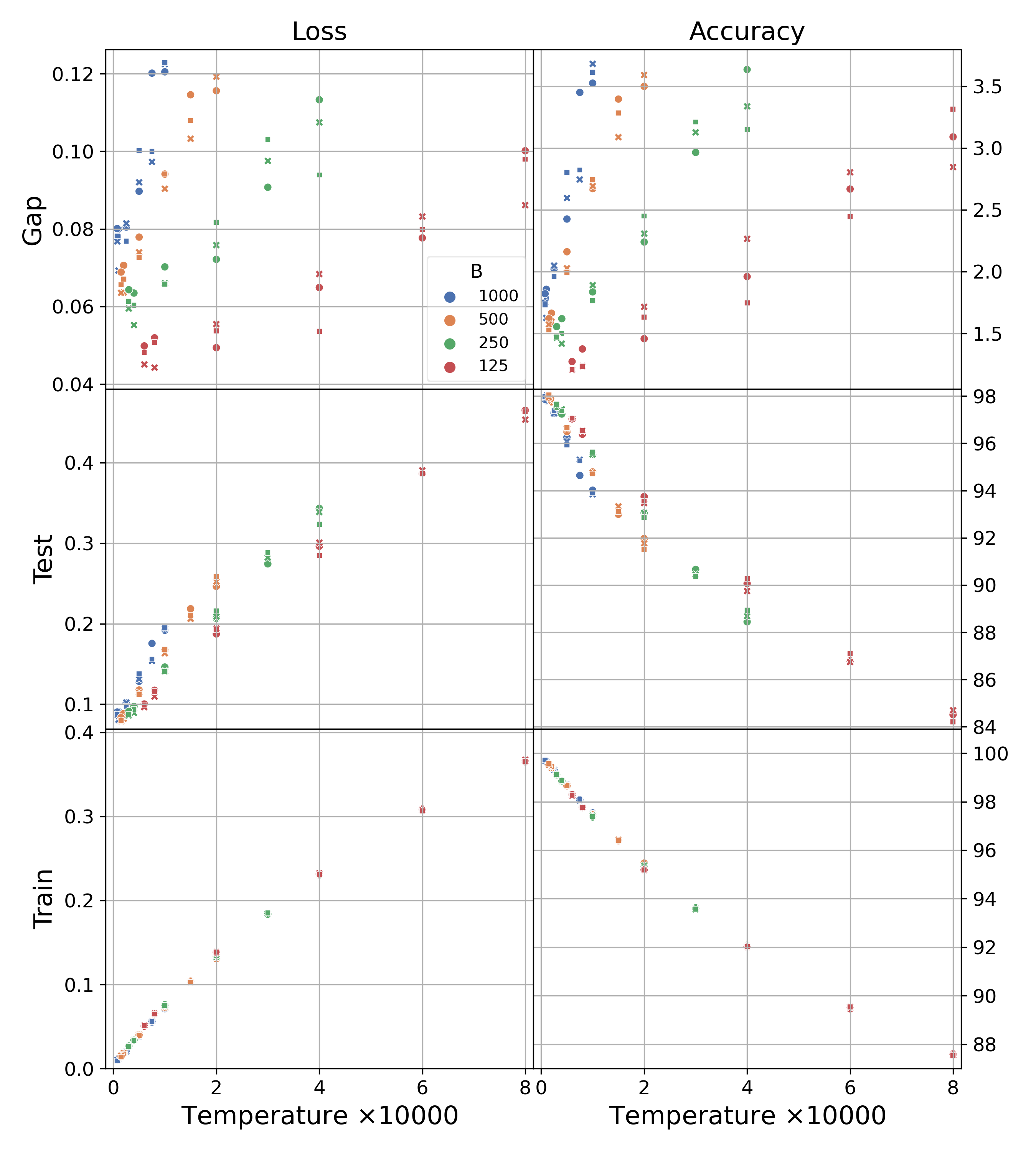}
\caption{Same setup as Fig. \ref{fig:PlainSGDK32DataAveraging} but using SGD with momentum. Different batch sizes lead to different curves for the gap and test metrics. All metrics here worsen as $T$ increases.}
\label{fig:SGDMomentumK32DataAveraging}
\end{figure}

We repeated the experiments in our earlier plots for the four variants of SGD when the data sets are sampled randomly without replacement and independently from the 60k Cifar10 data set. Statistically, the resulting sampling of the training set is identical to the data splitting one we have discussed. But the test set is very different because it now has a high percentage of data points that are also in the train set, although the specific number varies from sample to sample. Figures \ref{fig:PlainSGDK32DataAveraging}, \ref{fig:SGDCosineK32DataAveraging}, \ref{fig:SGDCosineMomentumK32DataAveraging}, and \ref{fig:SGDMomentumK32DataAveraging} show the results which show a much higher variance across experiments for each $T,$ but are otherwise qualitatively analogous to those produced with data splitting as a way to sample data sets. That is, the mean over the different experiments for each $T$ follows Eq. \ref{eq:GapVsT} for the gap, shows a positive finite temperature that optimizes test performance (although even higher temperatures seem needed to get there when there is no momentum but there is cosine learning schedule), and is non-decreasing for train performance for all SDG variants except SGD with momentum and no learning schedule.

\end{document}